\documentclass[10pt,letterpaper]{article}
\usepackage[T1]{fontenc}
\usepackage[letterpaper,left=1.05in,right=1.05in,top=.70in,bottom=.77in,footskip=23pt]{geometry}
\usepackage{amsmath,amssymb}
\usepackage{newtxtext,newtxmath}
\usepackage[scaled=.94]{helvet}
\usepackage{microtype}
\usepackage[dvipsnames]{xcolor}
\usepackage{graphicx,booktabs,multirow,tabularx}
\usepackage{xspace,soul}
\usepackage[normalem]{ulem}
\usepackage[shortlabels,inline]{enumitem}
\usepackage{tikz}
\usepackage{tcolorbox,titlesec,fancyhdr}
\usepackage[numbers,sort&compress]{natbib}
\usepackage[font=small,labelfont=bf,labelsep=period]{caption}
\usepackage{subcaption}
\definecolor{ink}{HTML}{151B2B}
\definecolor{linkblue}{HTML}{314D9A}
\definecolor{quiet}{HTML}{515969}
\definecolor{rulegray}{HTML}{CBD1DB}
\definecolor{coverink}{HTML}{1C2B34}
\definecolor{abstractfill}{HTML}{F1F4F6}
\usepackage[breaklinks,colorlinks,allcolors=linkblue]{hyperref}
\usepackage[capitalize]{cleveref}
\crefname{section}{Sec.}{Secs.}
\Crefname{section}{Section}{Sections}
\crefname{table}{Tab.}{Tabs.}
\Crefname{table}{Table}{Tables}
\IfFileExists{T1SourceSansThree-TLF.fd}
  {\newcommand{\coversans}{\fontfamily{SourceSansThree-TLF}\selectfont}}
  {\newcommand{\coversans}{\fontfamily{SourceSansPro-TLF}\selectfont}}
\titleformat{\section}{\Large\sffamily\bfseries\color{ink}}{\color{linkblue}\thesection}{.65em}{}
\titleformat{\subsection}{\large\sffamily\bfseries\color{ink}}{\thesubsection}{.65em}{}
\titleformat{\subsubsection}{\normalsize\sffamily\bfseries\color{ink}}{\thesubsubsection}{.65em}{}
\titleformat{\paragraph}[runin]{\normalsize\sffamily\bfseries\color{ink}}{\theparagraph}{.5em}{}
\titlespacing*{\section}{0pt}{2.1ex plus .6ex minus .3ex}{.65ex}
\titlespacing*{\subsection}{0pt}{1.65ex plus .4ex minus .25ex}{.45ex}
\renewenvironment{abstract}{%
  \begin{tcolorbox}[width=\linewidth,colframe=abstractfill,colback=abstractfill,
    boxrule=0pt,arc=6pt,outer arc=6pt,boxsep=0pt,left=13pt,right=13pt,
    top=10pt,bottom=11pt,before skip=0pt,after skip=0pt]
  {\coversans\fontseries{sb}\fontsize{8.1}{10}\selectfont\color{quiet}%
   \textls[130]{ABSTRACT}\par}
  \vspace{4.5pt}
  \fontsize{10.3}{12.8}\selectfont\color{coverink}%
  \setlength{\parindent}{0pt}\noindent\ignorespaces
}{\par\end{tcolorbox}}
\fancypagestyle{cover}{\fancyhf{}\fancyfoot[C]{\small\color{quiet}\thepage}}

\newcommand{\papertitle}{OLED: Online Latent Exception Detector via Semantic Rules and Jev for Embodied Tasks}

\usepackage{array,tabularx}
\newcolumntype{Y}{>{\raggedright\arraybackslash}X}

\usepackage{algorithm}
\usepackage{algpseudocode}

\newcommand{\RuleSet}{\mathcal{R}}
\newcommand{\GoalSet}{\mathcal{G}}

\DeclareMathOperator{\acc}{acc}
\newcommand{\ind}[1]{\mathbf{1}\!\left\{#1\right\}}

\usepackage{placeins}

\title{\papertitle}
\date{}
\hypersetup{pdftitle={\papertitle},
  pdfauthor={Xueliang Sun, Ziyi Liu, Bo Liu, Wanyou Wang, Qichen Tan, Jiajie Xu, He Huang, Min Zhang, Zhijun Li},
  pdfdisplaydoctitle=true,bookmarksnumbered=true}

\newcommand{\projectpageurl}{https://xinshangqiu.github.io/OLED-Project/}
\newcommand{\codeurl}{https://github.com/xinshangqiu/OLED}

\author{ 
\mbox{Xueliang Sun\textsuperscript{1}},
\mbox{Ziyi Liu\textsuperscript{2,\textdagger}},
\mbox{Bo Liu\textsuperscript{1}},
\mbox{Wanyou Wang\textsuperscript{1}},
\mbox{Qichen Tan\textsuperscript{2}},
\mbox{Jiajie Xu\textsuperscript{2}},
\mbox{He Huang\textsuperscript{2}},
\mbox{Min Zhang\textsuperscript{1}},
\mbox{Zhijun Li\textsuperscript{1,\textdagger}}
}

\newcommand{\mainpaperref}[1]{\ref{#1}}
\newcommand{\mainpapercref}[1]{\cref{#1}}

\begin{document}
\thispagestyle{cover}
\noindent\includegraphics[height=23.5pt]{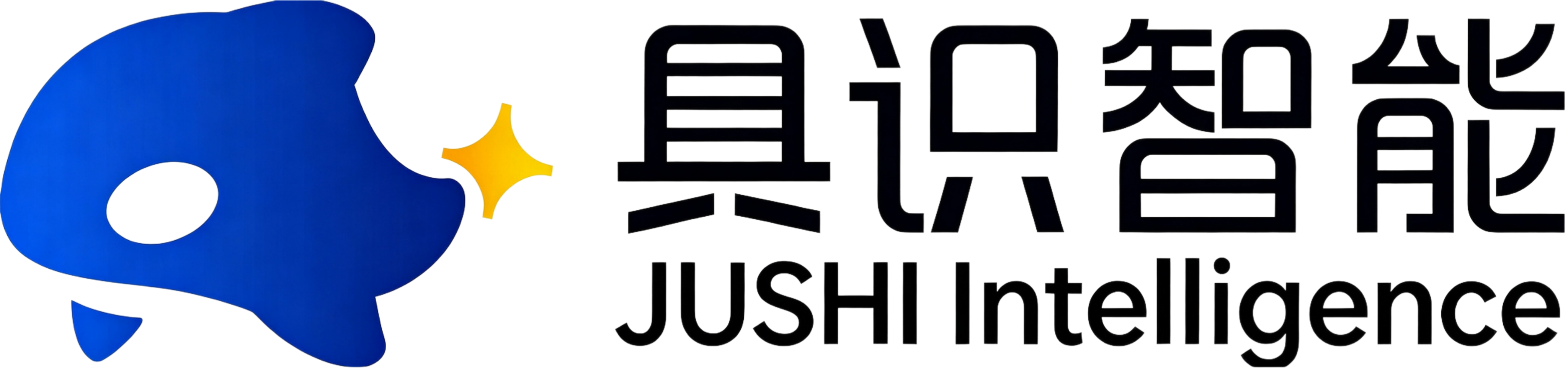}\par
\vspace{16pt}
{\noindent\raggedright\hyphenpenalty=10000\coversans\fontseries{sb}%
 \fontsize{18.05}{21.5}\selectfont\color{coverink}\papertitle\par}
\vspace{7pt}{\color{rulegray}\hrule height .45pt}
\vspace{8pt}
\makeatletter
{\noindent\raggedright\coversans\fontseries{sb}\fontsize{9.6}{12}\selectfont%
 \color{coverink}\@author\par}
\makeatother
\vspace{4pt}
{\noindent\fontsize{9.25}{11.5}\selectfont\color{quiet}%
 \textsuperscript{1}\,\raisebox{-.6pt}{\includegraphics[height=7.2pt]{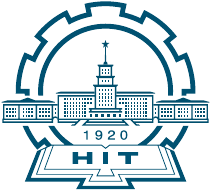}}%
 \hspace{.28em}Harbin Institute of Technology, Harbin, Heilongjiang, China.\par
 \noindent\textsuperscript{2}\,\raisebox{-.6pt}{\includegraphics[height=7.2pt]{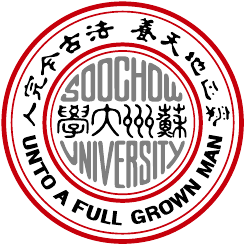}}%
 \hspace{.28em}Soochow University, Suzhou, Jiangsu, China.\par}
\vspace{3.5pt}
{\noindent\fontsize{9}{11}\selectfont\color{quiet}%
 First author\enspace
 \href{mailto:23B903018@stu.hit.edu.cn}{\nolinkurl{23B903018@stu.hit.edu.cn}} (Xueliang Sun)\par
 \noindent\textsuperscript{\textdagger}Corresponding authors\enspace
 \href{mailto:liuziyi@suda.edu.cn}{\nolinkurl{liuziyi@suda.edu.cn}} (Ziyi Liu)\quad
 \href{mailto:lizhijun_os@hit.edu.cn}{\nolinkurl{lizhijun_os@hit.edu.cn}} (Zhijun Li)\par}
\nointerlineskip
\vbox to 23pt{\vfil
  \hbox to \linewidth{%
    {\fontsize{10.5}{13}\selectfont\hypersetup{urlcolor=linkblue}%
    \href{\projectpageurl}{%
      \raisebox{-.65pt}{\includegraphics[height=9.25pt]{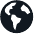}}%
      \hspace{3pt}Project Page}%
    \hspace{11pt}\textcolor{coverink}{\textbar}\hspace{11pt}%
    \href{\codeurl}{%
      \raisebox{-.65pt}{\includegraphics[height=9.25pt]{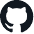}}%
      \hspace{3pt}Code}}%
    \hfil}%
  \vfil}

\begin{abstract}
Failures in embodied task execution often originate from exceptions that occur in earlier steps but remain undetected.
By the time the resulting task failure becomes apparent, recovery may no longer be possible.
Detecting such exceptions on unseen tasks requires distinguishing task-relevant violations from legitimate intermediate behavior.
To address this challenge, we propose OLED (Online Latent Exception Detector), a dual-model framework for latent exception detection. 
OLED combines an online Jev-based judgment model for timely exception detection with an offline model that generates task-specific semantic rules from reusable source-task priors.
A temporal controller integrates goal-related and event evidence across monitoring windows, deferring ambiguous goal-related exceptions according to manipulation context. 
The framework requires neither access to policy-internal signals nor risk-classifier training on target-task data.
Across four simulation and real-robot datasets with task-disjoint evaluation, OLED improves balanced accuracy over ViFailback-8B by 3.1–10.3 percentage points.
On ViFailBack, it increases strict ±25-frame window accuracy from 23.75\% to 38.75\%.
\end{abstract}
\par\vspace{12pt}
\section{Introduction}
\label{sec:intro}

\begin{figure}[!htbp]
    \centering
    \includegraphics[width=\linewidth]{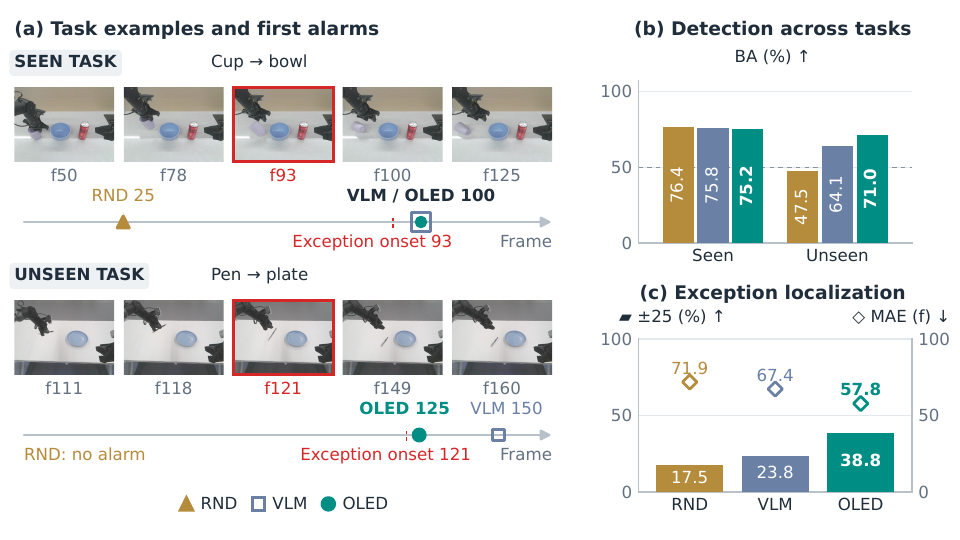}
    \caption{\textbf{Latent exception detection on ViFailBack.}
    (a) Task examples and timelines show exception onset and first alarms in seen and unseen tasks.
    (b) Balanced accuracy~(BA) across task groups.
    (c) Onset localization accuracy within $\pm25$ frames (bars) and mean absolute error (MAE) over detected failures (diamonds).
    VLM denotes ViFailback-8B.}
    \label{fig:teaser} 
\end{figure}

Embodied tasks involve a sequence of interactions with objects and their surroundings~\cite{Capx,mees2022calvin,liu2023libero}.
An exception in an early step can affect subsequent actions before the task is recognized as failed.
Consider the pen-to-plate task in Fig.~\ref{fig:teaser}.
The pen slips from the gripper during transport, but the robot continues executing the task.
The slip marks the onset of an exception, while the task failure becomes apparent only later.
Detecting the problem after execution leaves little opportunity for recovery.
This is particularly important for \textit{Vision-Language-Action (VLA)} policies~\cite{openvla, pi05} that perform multi-step manipulation tasks.
Reliable monitoring must therefore identify exceptions before they lead to task failure, while distinguishing them from normal intermediate behavior.
In this work, we study \emph{latent exception detection}, focusing on identifying task-relevant exceptions and locating their onset during execution.

Existing methods monitor robotic execution through policy signals or visual observations.
Some approaches detect failures by identifying deviations from successful execution patterns~\cite{xu2025faildetect,zhou2026rcnf}.
Such deviations can also occur during valid executions of unseen tasks, making them unreliable indicators of task-relevant exceptions.
Learned monitors, such as SAFE~\cite{Gu2025SAFE}, use policy-internal features and labeled rollouts to predict failures, but depend on access to the deployed policy and suitable training data.
Vision-language approaches instead interpret observations using task descriptions and constraints~\cite{guo2024doremi,zhou2025cam}.
These methods provide a flexible interface for monitoring different tasks, but reliable detection still requires understanding which conditions should hold and when they apply~\cite{zhao2026flare}.
Furthermore, predicting task failure does not necessarily identify the earlier exception that caused it.
Our work focuses on detecting these exceptions as execution unfolds, without relying on policy-internal signals or task-specific failure training.

Semantic constraints provide a way to describe what should happen during task execution~\cite{guo2024doremi,zhou2025cam}.
Beyond geometric properties, they capture object roles, task goals, and conditions that should hold during different interactions~\cite{ji2025robobrain,majumdar2024openeqa}.
In the pen-to-plate task, for example, the monitor must recognize the pen as the object being transferred and determine whether it remains held during transport.
These conditions can be expressed as questions about visual observations.
Jev~\cite{deusser2026jev} is a non-generative decision model that produces categorical judgments and associated probabilities without generating free-form text.
Its image-input interface~\cite{typesafe2026jevvision} enables such questions to be evaluated directly from visual observations.

Building on this capability, we propose \textit{Online Latent Exception Detector (OLED)}, a dual-model framework that combines offline semantic rule generation and retrieval with online Jev-based visual assessment and temporal risk inference. Applying this framework to multi-step robotic execution involves three key challenges.

\noindent\uline{\textbf{Incomplete semantic coverage.}}
Task instructions rarely specify all conditions needed to detect exceptions.
Generating monitoring questions directly from an instruction can overlook important object relations, violation conditions, and legitimate intermediate states.
These omissions can lead to missed exceptions or false positives, especially on unseen tasks.
To address this problem, we use an offline language model to generate reusable semantic rule families from source-task specifications.
Each family describes when a constraint applies, what constitutes an observable violation, and which intermediate states are acceptable.
For a new task, the rule generator adapts reusable rule
families using its specification and initial RGB observation.
This provides task-specific monitoring criteria without collecting new failure rollouts or training a separate risk classifier.

\noindent\uline{\textbf{Context-dependent assessment.}}
The same observed state can be normal or problematic depending on the current execution context.
For example, a pen resting on the table is normal before grasping but indicates a potential exception if it has been dropped during transport.
Without explicit execution-stage labels, the monitor must infer when each constraint applies from observed actions and task progress.
We address this challenge with an online Jev-based judgment model that jointly assesses event violations, goal satisfaction, goal-specific manipulation, and task engagement.
The model uses recent RGB observations, earlier reference frames, and relative motion cues to interpret these conditions in context.
Their combined judgments distinguish valid intermediate behavior from potential exceptions. All judgments are collected in a single model call per monitoring window, reducing online inference overhead.

\noindent\uline{\textbf{Uncertain temporal evidence.}}
Individual visual judgments can be noisy or incomplete, making it difficult to distinguish temporary deviations from actual exceptions.
An unmet goal can reflect ongoing manipulation, while a confirmed violation can remain relevant after its visual evidence disappears.
These situations require different forms of temporal reasoning.
We introduce a temporal inference module that combines goal-related evidence with event and motion candidates.
For goal-related risks, it accumulates evidence across observation windows while accounting for manipulation activity and uncertainty.
Contextual confirmation determines when to report candidates, deferring ambiguous goal-related decisions and requiring confirmation across consecutive queries for event candidates.
The right-hand panels of \cref{fig:teaser} compare trajectory-level detection across seen and unseen tasks and temporal localization.
OLED improves unseen-task balanced accuracy and strict $\pm25$-frame accuracy over ViFailback-8B.

Our main contributions are summarized as follows.
\begin{itemize}[leftmargin=0.6cm]
    \setlength{\itemsep}{2pt}
    \setlength{\topsep}{3pt}
    \setlength{\parsep}{0pt}
    \item We study latent exception detection in embodied tasks, focusing on identifying task-relevant exceptions before task failure and locating their onset during execution.

    \item We propose a dual-model framework that combines transferable semantic rules with efficient Jev-based visual assessment and temporal risk inference.
    By separating goal confirmation from event memory, the framework distinguishes legitimate intermediate behavior from persistent exceptions.

    \item We evaluate detection on four simulation and real-robot cohorts with held-out tasks, and temporal localization
    against annotated reference frames on ViFailBack, demonstrating effective exception detection and accurate onset localization across diverse embodied tasks.
\end{itemize}

\FloatBarrier

\section{Related Work}
\label{sec:related_work}

\subsection{Failure Detection from Policy Signals}
\label{sec:rw-policy-signals}

Policy-signal monitors estimate execution risk from policy inputs, representations, or actions~\cite{liu2024modelbased,wong2022erroraware,hoque2022thriftydagger,farid2022failure,zheng2026rewindil}.
One line of work identifies departures from successful execution without requiring failure examples.
FAIL-Detect~\cite{xu2025faildetect} formulates runtime monitoring through uncertainty-aware out-of-distribution detection, while FIPER~\cite{Romer2025FIPER} combines observation novelty in the policy's embedding space with uncertainty in generated action chunks.
Both use successful executions to calibrate their detection criteria.
A complementary approach learns failure predictors from labeled rollouts: SAFE~\cite{Gu2025SAFE} uses VLA internal features to estimate task-failure likelihood and evaluates transfer to unseen tasks.
Hybrid monitors combine policy evidence with semantic assessments.
Sentinel~\cite{Agia2025Sentinel} checks temporal action consistency for erratic behavior and uses VLM-based progress assessment for executions that remain consistent but do not solve the task.
These approaches provide useful evidence when the required policy interfaces and reference rollouts are available.
OLED targets an observation-based setting in which policy-internal representations and action outputs are unavailable.
It uses task specifications and external RGB observations to assess semantic violations, with monitoring parameters selected on source tasks rather than a risk classifier trained on target-task failures.

\subsection{Vision-Language-Model Failure Monitoring}
\label{sec:rw-vision-language}

Vision-language methods interpret execution in relation to task descriptions, supporting failure detection, diagnosis, and explanation~\cite{hosseinzadeh2026kite,ye2026robofac,sinha2024aesop}.
Existing approaches learn from generated or annotated failures~\cite{duan2025aha,ifailsense,vifailback,pacaud2026guardian,qi2026armor}, retrieve failure memories~\cite{ying2026robofailring}, or compare execution with nominal demonstrations~\cite{rolland2026fidel}.
These strategies supply different forms of reference knowledge: failure examples describe what can go wrong, memories provide related incidents, and demonstrations specify expected execution.
For online monitoring on a new task, the detector must also determine whether an observed deviation is relevant to the current instruction and whether it is compatible with a legitimate intermediate state.
Structured evidence and task constraints make these judgments more explicit~\cite{huang2025rekep,huang2023innermonologue,singh2023progprompt}.
REFLECT~\cite{liu2023reflect} organizes multisensory observations into hierarchical summaries for failure explanation and correction.
DoReMi~\cite{guo2024doremi} checks skill-specific constraints through visual question answering, while Code-as-Monitor~\cite{zhou2025cam} represents spatio-temporal constraints using geometric elements and generated monitoring code.
OLED builds on semantic constraint monitoring by transferring reusable rule families to new tasks.
Each instantiated rule specifies applicability, observable violation evidence, and acceptable exceptions.
This representation supports task-specific judgments without explicit execution-stage labels.

\subsection{Structured Decision Models for Detection}
\label{sec:rw-structured-decisions}

Structured decision models return predefined judgments and associated probabilities that downstream monitoring logic can consume directly.
Jev provides typed decisions without free-form text generation and supports multiple questions over shared image observations~\cite{typesafe2026jev,typesafe2026images}.
Such outputs avoid the need to recover categorical evidence from narrative responses, although a structured output alone does not establish when an alarm should be issued.
Prior applications use Jev for defect classification~\cite{megahed2026beyondclip}, while RoboJev~\cite{chen2026robojev} fine-tunes JEV-27B-VL for episode-level task-success assessment from sampled execution frames.
Online exception monitoring requires reasoning about observations before the final task outcome is known.
An unmet goal may indicate ongoing manipulation, whereas an observed adverse event can remain relevant after it leaves the current view.
OLED therefore uses frozen Jev as an evidence assessor: one call per monitoring window jointly evaluates event violations, goal relations, and manipulation activity.
The temporal module retains uncertain judgments, accumulates goal-related evidence, and maintains event memory before applying contextual confirmation.
This separates visual assessment from alarm timing and supports detecting exceptions during execution without fine-tuning the visual assessor on target tasks.

\FloatBarrier
\begin{figure}[!htbp]
  \centering
  \includegraphics[width=\linewidth]{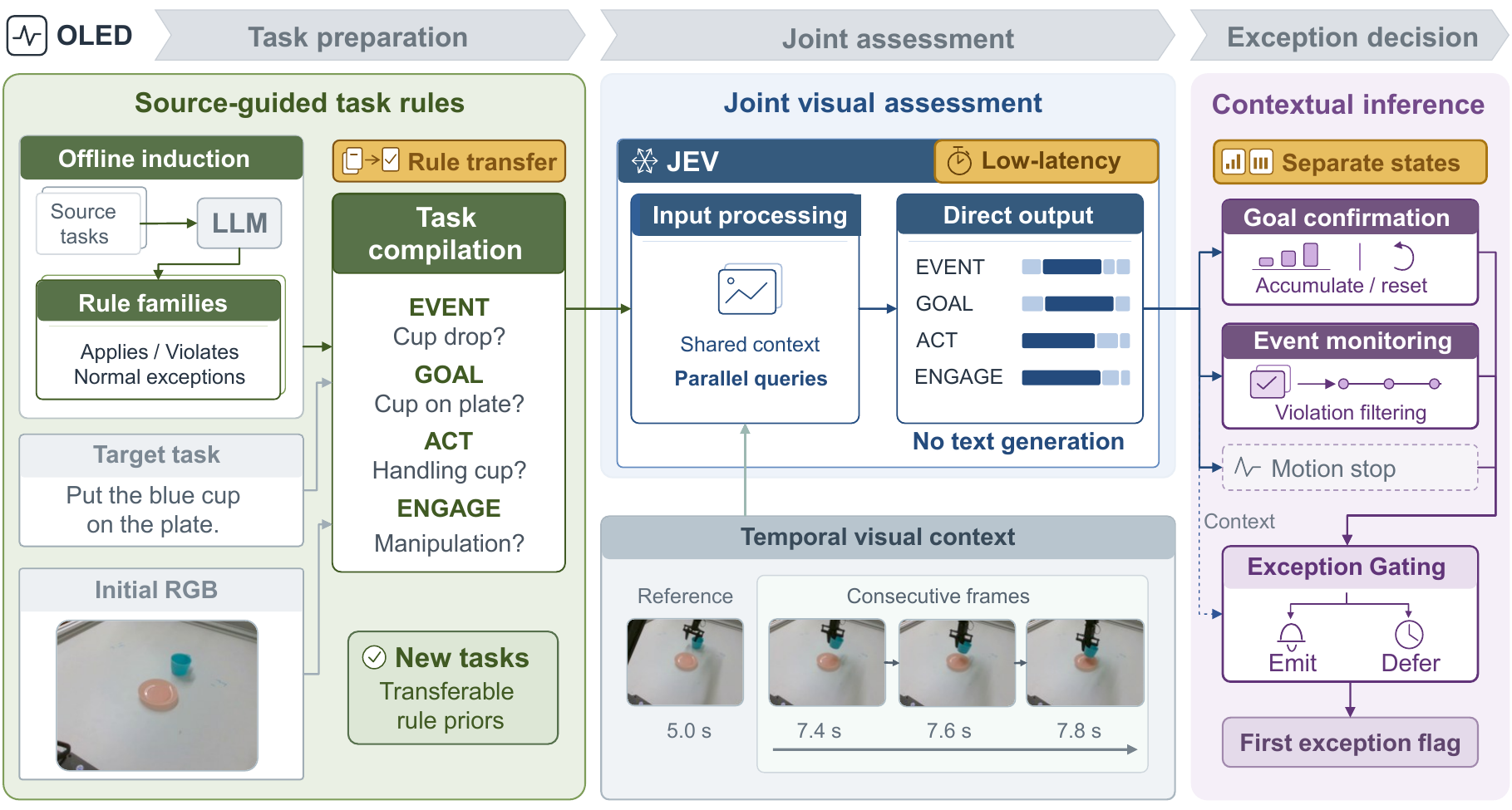}
  \caption{Overview of our framework for failure detection on new tasks, combining transferable rules, one joint Jev assessment per window,
and temporal risk inference.}
  \label{fig:pipeline}
\end{figure}

\section{Problem Formulation}
\label{sec:setup}

\textbf{Task Specification} $z$ contains a language instruction, relevant objects, and required goal relations $\GoalSet(z)$.
Robot execution produces an RGB observation sequence
$X=(x_0,\ldots,x_T)$. At time $t\in\mathcal Q\subseteq\{0,\ldots, T\} $, the monitor uses only the task specification and observations available up to $t$, without access to policy-internal states, action outputs, or execution-stage labels.

\noindent\textbf{Execution Exception} is a task-relevant adverse event or a persistent lack of goal progress that is inconsistent with normal task execution. Such exceptions do not necessarily cause immediate task failure, and their consequences can become apparent later.

\noindent\textbf{Online Latent Exception Detection.}
Given a task specification $z$ and streaming RGB observations,
we seek to develop an online detector $\mathcal D$ that identifies
task-relevant exceptions close to their onset on unseen tasks,
while avoiding false alarms during legitimate intermediate behavior.
At each monitoring time $t\in\mathcal Q$, the detector outputs
\begin{equation}
    R_t=\mathcal D(z,X_{\le t})\in\{0,1\},
    \label{eq:problem}
\end{equation}
where $X_{\le t}=(x_0,\ldots,x_t)$ and $R_t=1$
indicates an alarm.
The first alarm occurs at
$t_{\mathrm{det}}=\min\{t\in\mathcal Q:R_t=1\}$,
with $t_{\mathrm{det}}=\infty$ if no alarm is raised.

\FloatBarrier
\section{Method}
\label{sec:method}

As illustrated in \cref{fig:pipeline}, OLED combines  task rules, joint visual assessment, and temporal alarm inference.
An offline language model induces reusable rule families from static source-task specifications.
For a target task, instantiated rules are combined with goal relations and fixed templates to construct monitoring queries, with the initial RGB observation.
At each window, one Jev call evaluates these queries using available RGB observations and motion descriptors.
A separate temporal module integrates the resulting probabilities over time to generate exception candidates from goal, event, and auxiliary motion evidence.
Contextual Confirmation determines whether to emit or defer alarms based on manipulation evidence, without additional visual-model calls.

\begin{table}[t]
\centering
\small
\setlength{\tabcolsep}{3pt}
\renewcommand{\arraystretch}{1.05}
\caption{Representative rule families.
A, V, and E denote applicability, violation evidence, and legitimate exceptions, respectively.}
\label{tab:rule-semantics}
\begin{tabular}{@{}
  >{\raggedright\arraybackslash}p{0.20\linewidth}
  >{\raggedright\arraybackslash}p{0.60\linewidth}@{}}
\toprule
\textbf{Rule family} & \textbf{Reusable semantics} \\
\midrule
\textbf{Drop}
& \textbf{A:} Object $o$ is held and moved toward \(l\).\newline
  \textbf{V:} $o$ separates and falls before valid release.\newline
  \textbf{E:} Valid release; intermediate support. \\
\addlinespace[3pt]
\textbf{Wrong destination}
& \textbf{A:} The task specifies a destination \(l\)
for object \(o\). \newline
  \textbf{V:} $o$ is released at a different destination.\newline
  \textbf{E:} Task-valid intermediate placement. \\
\addlinespace[3pt]
\textbf{Topple requiring recovery}
& \textbf{A:} Stable object $o$ is approached or manipulated.\newline
  \textbf{V:} $o$ topples into a fallen pose requiring recovery.\newline
  \textbf{E:} Task-required tipping; $o$ remains graspable after minor tilt. \\
\bottomrule
\end{tabular}
\end{table}

\subsection{Source-Guided Task Rule Generation}
\label{sec:rules}

Task instructions specify desired outcomes but rarely enumerate the conditions that distinguish execution exceptions from legitimate intermediate behavior.
Generating monitoring constraints from target-task information alone may therefore overlook relevant violations or acceptable intermediate states.
We address this gap by transferring structured rule priors from source tasks and adapting them to the target task and its visual scene.

\paragraph{Structured rule priors.}
Let \(\mathcal{S}_m\) denote the source-task information collected from dataset \(m\), including task instructions, object inventories, and required goal relations.
An offline language model aggregates information from \(M\) source datasets to induce reusable rule families  \(\mathcal{F}\):
\begin{equation}
    \mathcal{S}_{\mathrm{src}}
    = \bigcup_{m=1}^{M}\mathcal{S}_m,
    \qquad
    \mathcal{F}
    = \operatorname{LLM}(\mathcal{S}_{\mathrm{src}}).
    \label{eq:induction}
\end{equation}
Held-out target tasks are excluded from this induction process.
Each family \(f\in\mathcal{F}\) is represented as \(f=(a_f,v_f,e_f)\), where \(a_f\) specifies applicability conditions, \(v_f\) describes observable violation evidence, and \(e_f\) identifies normal exceptions.
This representation captures both adverse-event criteria and the qualifications needed to distinguish them from legitimate execution.
For example, a delivery rule identifies placement at an incorrect destination as violation evidence while allowing transport toward the intended target.
These reusable semantic conditions, summarized in \cref{tab:rule-semantics}, provide the knowledge transferred to new tasks.

\paragraph{Task-specific rule generation.}
For a new task, a rule generator combines the source rule families \(\mathcal{F}\), task specification \(z\), and initial RGB observation \(x_0\) to produce monitoring constraints:
\begin{equation}
    \RuleSet(z;x_0)
    = \operatorname{Generate}(\mathcal{F},z,x_0).
    \label{eq:compile}
\end{equation}
Here, \(\RuleSet(z;x_0)\) denotes the generated event constraints, abbreviated as \(\RuleSet(z)\) below.
The task specification defines the relevant objects and required goal relations, while the RGB observation provides visual context for grounding these entities in the current scene.
Source priors guide the adaptation of applicability conditions, violation criteria, and normal exceptions into task-specific constraints.
The required goal relations \(\GoalSet(z)\) remain as specified by \(z\), and an unmet goal alone does not constitute an event violation.
The generated constraints define what evidence the monitor should assess, and fixed templates express them as visual queries for Jev.
These rules and queries remain fixed during execution, while subsequent observations determine whether their conditions are satisfied.

\subsection{Joint Visual Assessment}
\label{sec:joint}

Assessing task constraints requires evidence of both their current status and the surrounding manipulation context.
We therefore use a frozen Jev model to jointly assess event violations, goal relations, and activity from shared visual observations, producing structured evidence for temporal inference.

\paragraph{Temporal visual context.}
\label{sec:evidence}

At query index \(t\), we combine recent frames indexed by \(\mathcal{W}_t\) with earlier reference frames indexed by \(\mathcal{A}_t\):
\begin{equation}
        X_{\le t}^{\mathrm{sel}}
        = (x_s)_{s\in\mathcal{I}_t},   \qquad
        \mathcal{I}_t
        = \mathcal{W}_t\cup\mathcal{A}_t
        \subseteq \{0,\ldots,t\}.
    \label{eq:visual-evidence}
\end{equation}
Recent frames provide evidence of ongoing manipulation, current goal relations, and newly occurring events.
Earlier references supply context for identifying task-relevant objects and interpreting their relations in the scene.
All selected frames are presented chronologically and are available at the current query endpoint.
When history is limited, we shorten the recent sequence and set \(\mathcal{A}_t=\varnothing\) if no earlier references are available.

An optional motion descriptor \(\omega_t\) characterizes visual changes between successive sampled RGB frames.
We compute dense optical flow for each frame pair and subtract its component-wise spatial median to obtain residual motion~\cite{farneback2003}.
The descriptor compares recent residual motion with a reference derived solely from earlier observations, indicating whether frame-to-frame activity has increased or decreased.
This comparison provides context for assessing relative rest, but reduced activity alone does not establish an execution exception.

\paragraph{Joint semantic assessment.}
\label{sec:assessment}

The event, goal, and manipulation queries are summarized in \cref{tab:heads}.
EVENT queries assess whether a generated constraint applies and whether its violation condition is observed.
For each goal \(g\in\GoalSet(z)\), \(\mathrm{GOAL}_g\) assesses the required relation, while \(\mathrm{ACT}_g\) assesses goal-specific physical manipulation.
ENGAGE captures task-relevant manipulation at the execution level without implying that every goal has been attempted.
These distinctions allow the temporal module to interpret an unmet goal alongside evidence of ongoing manipulation, including activity associated with other goals.

A single Jev call evaluates these queries using the shared visual input and the motion descriptor when available:
\begin{equation}
    \mathbf{p}_t
    =
    \operatorname{JEV}\bigl(
        X_{\le t}^{\mathrm{sel}},z,
        \RuleSet(z),\GoalSet(z);\omega_t
    \bigr).
    \label{eq:joint}
\end{equation}
All assessments share the selected visual context, while the generated constraints and required goal relations specify the conditions to evaluate.
The probability collection \(\mathbf{p}_t\) includes both the semantic observations listed in \cref{tab:heads} and the auxiliary motion judgment.

\paragraph{Evidence acceptance.}

For query \(h\), let \(\mathcal{L}_h\) denote its admissible labels and \(p_{t,h}(\ell)\) the probability assigned to label \(\ell\).
For temporal decisions requiring explicit categorical evidence, a label is accepted only when it has maximal probability and reaches the corresponding threshold \(\tau_h\):
\begin{equation}
    \acc_{t,h}(\ell)
    =
    \ind{
        \begin{aligned}
            &p_{t,h}(\ell)\geq\tau_h\\
            &{}\land
            p_{t,h}(\ell)
            =
            \max_{\ell'\in\mathcal{L}_h}
            p_{t,h}(\ell')
        \end{aligned}
    }.
    \label{eq:accept}
\end{equation}
A goal remains unresolved when the model cannot confidently determine whether it is satisfied or unmet.
Likewise, the absence of accepted ACT \emph{yes} does not imply accepted ACT \emph{no}.
This distinction preserves uncertainty when the available observations do not support a definite judgment.
We retain the original probabilities alongside these acceptance predicates so that the temporal module can apply decision-specific evidence tests.

\begin{table}[t]
\centering
\small
\caption{Semantic queries for event, goal, and manipulation assessment.}
\label{tab:heads}
\begin{tabularx}{\linewidth}{@{}>{\raggedright\arraybackslash}p{0.19\linewidth}YY@{}}
\toprule
Output & Labels & Interpretation\\
\midrule
Event & violation, normal, not applicable, uncertain & Evidence of instantaneous or stage-related events\\
\(\mathrm{GOAL}_g\) & satisfied, unmet, uncertain & Current required relation for goal \(g\)\\
ENGAGE & yes, no, uncertain & Physical manipulation relevant to the task\\
\(\mathrm{ACT}_g\) & yes, no, uncertain & Goal-subject manipulation\\
\bottomrule
\end{tabularx}
\end{table}

\subsection{Temporal Exception Inference}
\label{sec:state}

An unmet goal may reflect ongoing manipulation, while a previously observed violation may remain relevant after it disappears from view.
We therefore combine goal monitoring \(U_t\) and event memory \(E_t\) to form an exception candidate \(B_t=U_t\lor E_t\), and use manipulation context to determine when to report it.
GOAL, ACT, and EVENT evidence follows \cref{eq:accept} with a shared threshold \(\tau\).

\paragraph{Goal monitoring.}
We monitor goal status across query windows to distinguish persistent lack of progress from legitimate intermediate behavior.
After task engagement is detected, we increment a goal-specific counter whenever the goal is accepted as unmet and every goal is either accepted as satisfied or supported by an accepted ACT \emph{no} judgment.
Once the counter reaches \(K\), the monitor generates an exception candidate for that goal.
Evidence of manipulation involving another goal resets this counter.
Consecutive \emph{uncertain} judgments for any goal pause accumulation without clearing the current count.

\paragraph{Event monitoring.}
Event and motion monitoring complement goal monitoring by capturing rule violations and possible interruptions in execution.
Event monitoring checks whether observed behavior violates task-specific constraints.
All event queries use the same acceptance criterion in \cref{eq:accept}, and a query generates a candidate only when its \emph{violation} label is accepted. 
Motion-stop monitoring examines whether previously active execution has entered a sustained period of rest before a required goal is achieved.
It generates a candidate when rest persists after task engagement and recent movement, while a goal remains unmet and no goal-related action is accepted as ongoing.
Both routes provide exception candidates for subsequent temporal confirmation.

\paragraph{Contextual confirmation.}
Contextual confirmation uses execution context to determine when to report an exception.
For goal candidates, let \(e_t\) denote the ENGAGE \emph{yes} probability and \(v_t\) the maximum EVENT \emph{violation} probability.
When no decision is pending, we defer a goal candidate if \(e_t\geq\theta_e\) and \(v_t<\theta_v\), since ongoing manipulation may explain why the goal remains unmet.
Otherwise, the candidate is reported immediately.
Once deferred, a goal decision is reported only when a subsequent query produces an exception candidate and ENGAGE is \emph{no}.
Similarly, event candidates require confirmation across consecutive query windows before an exception is reported.
A query discontinuity clears the pending decision before the current query is processed.
We record the first detection at the query where an exception
flag is issued. If no flag is issued during execution,
no detection time is reported.

\FloatBarrier

\section{Experiments}
\label{sec:experiments}

\subsection{Experimental Setup}
\label{sec:exp-setup}

\begin{table}[t]
\centering
\small
\setlength{\tabcolsep}{3.5pt}
\caption{Failure-detection performance (\%) over all scheduled queries
on four datasets, with the best BA in each dataset shown in bold.}
\label{tab:exp-main}

\begin{tabularx}{\textwidth}{@{}Y*{12}{r}@{}}
\toprule
Method
& \multicolumn{3}{c}{LIBERO-Pro}
& \multicolumn{3}{c}{ViFailBack}
& \multicolumn{3}{c}{SAFE-WidowX}
& \multicolumn{3}{c}{SAFE-Franka} \\
\cmidrule(lr){2-4}
\cmidrule(lr){5-7}
\cmidrule(lr){8-10}
\cmidrule(lr){11-13}
& BA $\uparrow$ & Rec. $\uparrow$ & FPR $\downarrow$
& BA $\uparrow$ & Rec. $\uparrow$ & FPR $\downarrow$
& BA $\uparrow$ & Rec. $\uparrow$ & FPR $\downarrow$
& BA $\uparrow$ & Rec. $\uparrow$ & FPR $\downarrow$ \\
\midrule

RND
& 55.33 & 100.00 & 89.33
& 47.52 & 26.15 & 31.11
& 49.87 & 41.55 & 41.80
& 48.12 & 35.13 & 38.89 \\

LogpZO
& 61.50 & 100.00 & 77.00
& 38.46 & 50.26 & 73.33
& 45.70 & 15.86 & 24.45
& 46.41 & 36.32 & 43.50 \\

Similarity
& 75.50 & 100.00 & 49.00
& 49.49 & 12.31 & 13.33
& 47.36 & 49.77 & 55.05
& 48.93 & 17.69 & 19.83 \\

SAFE-MLP
& 53.50 & 36.00 & 29.00
& 48.89 & 0.00 & 2.22
& 49.97 & 30.67 & 30.74
& 52.35 & 6.84 & 2.14 \\

SAFE-LSTM
& 53.83 & 24.00 & 16.33
& 48.89 & 6.67 & 8.89
& 53.08 & 15.86 & 9.70
& 51.50 & 16.15 & 13.16 \\

I-FailSense
& 54.50 & 76.00 & 67.00
& 50.00 & 100.00 & 100.00
& 50.21 & 11.34 & 10.93
& 48.97 & 15.13 & 17.18 \\

JEV direct
& 51.50 & 3.00 & 0.00
& 50.51 & 7.69 & 6.67
& 50.00 & 0.00 & 0.00
& 50.38 & 1.03 & 0.26 \\

ViFailback-8B
& 79.50 & 79.00 & 20.00
& 64.10 & 61.54 & 33.33
& 59.41 & 27.43 & 8.61
& 51.03 & 12.56 & 10.51 \\

\midrule
\textbf{OLED}
& \textbf{83.00} & 76.00 & 10.00
& \textbf{71.03} & 75.38 & 33.33
& \textbf{69.70} & 70.14 & 30.74
& \textbf{54.10} & 68.21 & 60.00 \\

\bottomrule
\end{tabularx}
\end{table}

\paragraph{Datasets.} 
We evaluate on four datasets spanning simulation and real-world manipulation.
LIBERO-Pro~\cite{liberopro} comprises 200 simulated $\pi_{0.5}$~\cite{pi05} rollouts across ten task families under structured perturbations, with ten successes and ten failures per family.
ViFailBack~\cite{vifailback} contains 80 real-world ALOHA trajectories from five held-out tasks, with 420 trajectories from 17 separate tasks forming the source pool.
SAFE-WidowX and SAFE-Franka~\cite{Gu2025SAFE} contain real-robot
executions of OpenVLA~\cite{openvla} and $\pi_0$-FAST-DROID~\cite{pertsch2025fast}, respectively.
WidowX covers 532 trajectories across eight lifting and pick-and-place tasks.
Our fixed Franka comprises 13 tasks spanning object placement, articulated-object manipulation, and other skills, with 30 successes and 30 failures per task.

\paragraph{Evaluation protocol and metrics.}
We use task-disjoint splits: ten-fold leave-one-task-out on LIBERO-Pro, 17 source and five held-out tasks on ViFailBack, and four-fold grouped splits on SAFE-WidowX and SAFE-Franka, with the remaining groups as sources.
To account for class imbalance, our primary metric is balanced accuracy~(BA), \(\mathrm{BA}=\frac{1}{2}(\mathrm{Recall}+1-\mathrm{FPR})\), alongside failure recall~(Rec.) and the false-positive rate~(FPR) on successful trajectories.
Where temporal annotations are available, strict \(\pm25\)-frame accuracy~($A_{25}$) counts successful trajectories with no predicted exception and failed trajectories whose predicted exception frame lies within 25 frames of the reference as correct.
Mean absolute error~(MAE) measures the average absolute difference between predicted exception frames and reference frames over method-specific detected failed trajectories.
P50 and P95 denote the 50th and 95th percentiles
of request latency, respectively.
All metrics are averaged over three runs.

\paragraph{Baselines and implementation.}
We compare our method with RND, LogpZO, Similarity, SAFE-MLP, SAFE-LSTM, I-FailSense~\cite{ifailsense}, ViFailback-8B, and direct Jev assessment.
We adapt detectors originally based on policy signals to RGB inputs using features extracted by a frozen DINOv2 encoder~\cite{dinov2}.
All methods are evaluated on matched cohorts using observations available at each query, with method-specific query schedules, frame counts, encoders, and calibration.
Our goal monitor requires \(K=4\) supporting queries to generate an exception candidate.
Parameter selection, calibration, and validation are performed exclusively on source-domain data.
Local models are trained and evaluated on an NVIDIA RTX 4090
GPU.
LIBERO-Pro uses 13 recent frames over 3\,s plus references 4 and 6\,s earlier; ViFailBack uses 13 over 75 source frames plus references 100 and 150 frames earlier.
Both SAFE datasets use five frames over 4\,s without additional references. Inputs are shortened when history is limited.
We use DeepSeek for rule-family induction and task compilation, and access Jev through TypeSafe's official API for online visual assessment.

\subsection{Comparison with Existing Detectors}
\label{sec:exp-comparison}

As shown in \cref{tab:exp-main}, OLED achieves the highest BA among the compared detectors on all four datasets.
Direct Jev assessment yields BA close to \(50\%\) on the three real-world datasets, with failure recall no higher than \(7.69\%\).
OLED detects more failures, although its balance between recall and false positives varies across datasets.

On LIBERO-Pro, OLED improves BA from \(79.50\%\) to \(83.00\%\) relative to ViFailback-8B.
This improvement comes from reducing FPR, despite a three-percentage-point decrease in recall.
On ViFailBack, OLED increases recall from \(61.54\%\) to \(75.38\%\) at the same FPR, improving failure coverage without additional false positives.
On SAFE-WidowX, the improvement involves a different trade-off: recall increases by \(42.71\) percentage points relative to ViFailback-8B, while FPR increases by \(22.13\) points.

SAFE-Franka exposes the main limitation of this balance.
OLED achieves \(68.21\%\) recall, but its \(60.00\%\) FPR means that more than half of successful executions are incorrectly classified as failures.
Consequently, its BA reaches only \(54.10\%\), exceeding the strongest baseline, SAFE-MLP, by \(1.75\) percentage points.
These results indicate that improved failure coverage does not yet translate into reliable false-positive control across all evaluated settings.

\subsection{Temporal Localization of Exceptions}
\label{sec:exp_localization}

\begin{figure}[!htbp]
  \centering
  \captionsetup{skip=3pt,justification=raggedright,singlelinecheck=false}
  \begin{minipage}[t]{0.48125\linewidth}
    \vspace{0pt} 
    \centering
    \includegraphics[width=\linewidth]{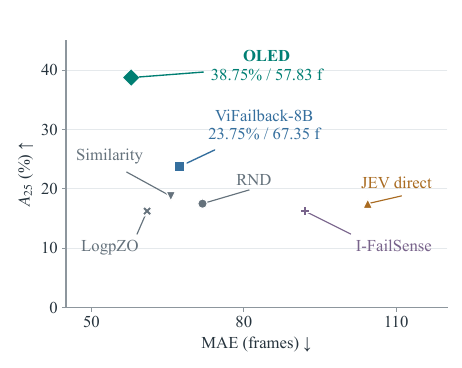}
    \caption{\textbf{First-detection localization on ViFailBack.} Higher $A_{25}$ and lower MAE are better; MAE is conditional on each method's detected failures.}
    \label{fig:exp_localization}
  \end{minipage}\hfill
  \begin{minipage}[t]{0.48125\linewidth}
    \vspace{0pt}
    \centering
    \includegraphics[width=\linewidth]{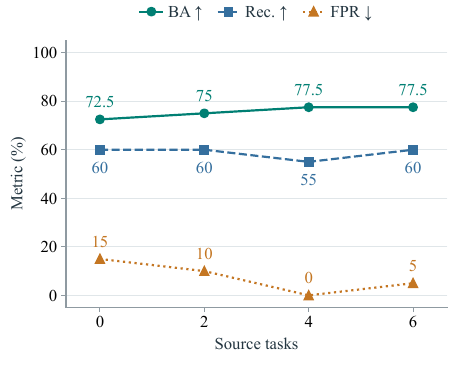}
    \caption{\textbf{Source task composition ablation.} All settings use the same 40 target trajectories. Zero means direct constraint generation without source priors.}
    \label{fig:source_rule_families}
  \end{minipage}
\end{figure}

We evaluate temporal agreement between predicted exception frames and annotated reference frames on ViFailBack.
As shown in \cref{fig:exp_localization}, OLED achieves the highest \(A_{25}\) and the lowest reported MAE among the compared methods.
Relative to ViFailback-8B, OLED improves \(A_{25}\) from \(23.75\%\) to \(38.75\%\) and reduces MAE from \(67.35\) to \(57.83\) frames.
The higher \(A_{25}\) indicates that more trajectories satisfy the strict evaluation criterion, which rewards both successful executions without exception predictions and failed executions localized within the reference window.
The lower reported MAE provides complementary evidence of closer temporal alignment among detected failures. 
LogpZO illustrates why both metrics are necessary: its MAE of \(60.96\) frames is relatively close to OLED's, but its \(A_{25}\) is only \(16.25\%\).
A low timing error among detected failures therefore does not necessarily imply accurate detection and localization across the full cohort.

\subsection{Ablation of Source Task Composition}
\label{sec:exp_source_task_diversity}

We compare direct constraint generation without source priors (zero source tasks) with generation guided by rule families induced from 2, 4, or 6 source tasks.
All settings use the same source data for calibration and are evaluated on two held-out LIBERO-Pro tasks comprising 20 successful and 20 failed trajectories.

As shown in \cref{fig:source_rule_families}, source priors improve BA by up to 5.00 percentage points, with both four and six source tasks reaching 77.50\%.
With six tasks, recall remains at 60.00\%, while FPR falls from 15.00\% to 5.00\%, corresponding to two fewer false positives.
The four-task setting eliminates all three baseline false positives but misses one additional failure, yielding the same BA.
These results suggest that source priors improve false-positive control on this cohort, while adding source tasks can change the recall--FPR trade-off without further increasing BA.

\subsection{Qualitative Analysis}
\label{sec:exp-qualitative}

\begin{figure}[!htbp]
  \centering
  \includegraphics[width=\linewidth]{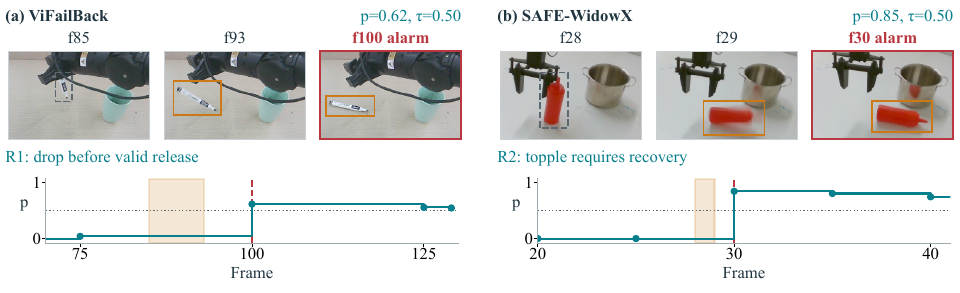}
  \caption{\textbf{Qualitative rule alignment.} (a) Marker drop on ViFailBack; (b) bottle toppling on SAFE-WidowX. Boxes and shaded intervals are manual correspondence aids. Dots show actual query probabilities, steps retain the latest output, dotted lines mark the threshold, and red dashed lines mark the first alarm. Post-alarm curves are retrospective replays.}
  \label{fig:qualitative-constraints}
\end{figure}

The two archived replays in \cref{fig:qualitative-constraints} illustrate the correspondence between physical transitions, task-bound constraints, and predicted exception frames.
Dots mark actual queries, while green steps retain the latest violation probability between queries.
Red dashed vertical lines mark predicted exception frames, while shaded regions indicate manually exception intervals containing adverse transitions.

In the  ViFailBack example, the marker separates from the gripper and falls onto the table before a valid return to the cup.
This matches \(R_1\), which identifies unintended drops while excluding valid placement and controlled release.
The first exception prediction occurs at frame 100, where the violation probability reaches \(0.62\), exceeding the threshold of \(0.50\).
In the SAFE-WidowX example, the bottle changes from an upright pose to a fallen position outside the pot.
This transition matches \(R_2\), which describes toppling that requires recovery before task completion.
The first exception prediction occurs at frame 30, with a violation probability of \(0.85\).
Together, these cases illustrate how task-bound constraints connect observable adverse transitions to explicit semantic evidence for exception detection.

\subsection{Ablation of Temporal Exception Inference}
\label{sec:exp-temporal}

We examine goal monitoring, event monitoring, and contextual confirmation through the ablations in \cref{tab:exp_temporal_ablation}.
Removing goal monitoring yields the largest BA decrease on all four datasets, supporting the importance of tracking persistent unmet goals.
Removing event monitoring lowers BA by 1.15--3.50 points, indicating that explicit violations complement goal-based monitoring.
On ViFailBack, removing confirmation reduces BA by 14.62 points, comparable to the 15.13-point decrease without goal monitoring.
This highlights the importance of both candidate generation and deciding when evidence warrants an exception report.
The smaller confirmation gains on the other datasets indicate that its benefit depends on the evaluation setting.

\begin{table}[t]
\centering
\small
\setlength{\tabcolsep}{6pt}
\caption{BA (\%) $\uparrow$ for temporal-inference ablations.}
\label{tab:exp_temporal_ablation}
\begin{tabular}{@{}lrrrr@{}}
\toprule
Variant & LIBERO-Pro & ViFailBack & WidowX & Franka \\
\midrule
w/o. goal-route  & 76.00 & 55.90 & 62.41 & 51.41 \\
w/o. event-route & 79.50 & 69.49 & 66.94 & 52.95 \\
w/o. confirmation & 80.00 & 56.41 & 67.30 & 53.21 \\
\midrule 
\textbf{OLED} & 83.00 & 71.03 & 69.70 & 54.10 \\
\bottomrule
\end{tabular}
\end{table}

\subsection{Alternative Visual Assessors and Monitors}
\label{sec:exp-assessor} 

We compare five API replacements for Jev within OLED and three detection-only DoReMi~\cite{guo2024doremi} variants in \cref{tab:exp-assessor}, with method-specific inputs and decision settings.
OLED achieves the highest $A_{25}$ (38.75\%), exceeding the best API and DoReMi variants by 5.00 and 3.75 percentage points.
At the same FPR, OLED localizes three additional failures within $\pm25$ frames compared with DoReMi-GPT despite lower BA and recall, highlighting the distinction between failure coverage and temporal agreement.
MAE remains conditional on each method's detected failures.

DoReMi has lower request P50 for all three backends, but improves BA only with GPT and reduces it with DeepSeek and Gemini.
OLED combines its leading $A_{25}$ with historical successful Jev request P50/P95 of 0.328/0.581\,s, below all API alternatives but above the local detectors.
Differences in workload and timing scope limit direct latency comparisons.

\begin{table}[tp]
  \centering
  \small
  \caption{Comparison of visual assessors and monitoring protocols on ViFailBack.
  BA, recall, FPR, and $A_{25}$ are percentages; MAE is in frames and P50/P95 are in seconds.}
  \label{tab:exp-assessor} 

  \setlength{\tabcolsep}{4pt}
  \begin{tabular}{@{}lrrrrr@{\hspace{8pt}}rr@{}}
    \toprule
    & \multicolumn{3}{c}{Detection} & \multicolumn{2}{c}{Localization} & \multicolumn{2}{c}{Latency} \\
    \cmidrule(lr){2-4}\cmidrule(lr){5-6}\cmidrule(lr){7-8}
    Model & BA $\uparrow$ & Rec.\ $\uparrow$ & FPR $\downarrow$ & $A_{25}$ $\uparrow$ & MAE $\downarrow$ & P50 $\downarrow$ & P95 $\downarrow$ \\
    \midrule
    \multicolumn{8}{@{}l}{\emph{Existing detectors}} \\
    ViFailback-8B & 64.10 & 61.54 & 33.33 & 23.75 & 67.35 & 0.188 & 0.209 \\
    I-FailSense & 50.00 & 100.00 & 100.00 & 16.25 & 92.00 & 0.243 & 0.319 \\
    \midrule
    \multicolumn{8}{@{}l}{\emph{OLED with API assessors}} \\
    GPT-6.1 Sol$^\dagger$ & 75.64 & 64.62 & 13.33 & 32.50 & 54.02 & 8.978 & 11.725 \\
    DeepSeek Flash$^\dagger$ & 67.95 & 69.23 & 33.33 & 32.50 & 65.40 & 1.725 & 3.154 \\
    GLM-5.3-Flash$^\dagger$ & 63.59 & 73.85 & 46.67 & 33.75 & 69.02 & 7.857 & 15.559 \\
    Grok-4.7$^\dagger$ & 54.10 & 81.54 & 73.33 & 26.25 & 76.62 & 7.222 & 17.201 \\
    Gemini-3-flash$^\dagger$ & 56.67 & 60.00 & 46.67 & 25.00 & 73.51 & 3.228 & 8.372 \\
    \midrule
    \multicolumn{8}{@{}l}{\emph{DoReMi-style monitoring (detection only)}} \\
    GPT-6.1 Sol$^\dagger$ & 78.72 & 90.77 & 33.33 & 35.00 & 45.53 & 6.177 & 11.628 \\
    DeepSeek Flash$^\dagger$ & 44.62 & 49.23 & 60.00 & 16.25 & 55.09 & 1.193 & 1.790 \\
    Gemini-3-flash$^\dagger$ & 51.79 & 56.92 & 53.33 & 16.25 & 76.70 & 2.888 & 5.822 \\
    \midrule
    \textbf{OLED} & \textbf{71.03} & 75.38 & 33.33 & \textbf{38.75} & 57.83 & 0.328 & 0.581 \\
    \bottomrule
    \addlinespace[2pt]
    \multicolumn{8}{@{}l}{\footnotesize $^\dagger$ Official API proxy.}
  \end{tabular}
\end{table}

\FloatBarrier
\section{Conclusion}
\label{sec:conclusion}

We presented OLED, a visual exception monitor for robot manipulation that combines transferable task rules, joint visual assessment, and contextual temporal inference.
Rule families induced from static source-task specifications adapt monitoring criteria to new tasks without training a new risk classifier.
A single Jev call per window jointly assesses violations, goals, and manipulation context, while temporal evidence and exception gating determine when to issue an alarm.
Experiments across four simulation and real-robot cohorts demonstrate the value of this design, with the highest balanced accuracy among the compared methods on all three real-world cohorts.
Ablations support the contributions of source rule priors and complementary temporal evidence.
Future work will focus on reducing false exception reports under visual ambiguity and evaluating the monitor within closed-loop recovery.

\FloatBarrier
{
    \small
    \bibliographystyle{oled-reference}
    \bibliography{references}
}

\clearpage
\phantomsection
\pdfbookmark[0]{Supplementary Material}{supplementary-material}
{\noindent\coversans\fontseries{sb}\large\color{coverink}\papertitle\par}
\vspace{6pt}
{\noindent\Large\sffamily\bfseries\color{ink}Supplementary Material\par}
\vspace{8pt}
\appendix
\setcounter{figure}{0}
\setcounter{table}{0}
\setcounter{equation}{0}
\setcounter{algorithm}{0}
\renewcommand{\theHfigure}{supp.\arabic{figure}}
\renewcommand{\theHtable}{supp.\arabic{table}}
\renewcommand{\theHequation}{supp.\arabic{equation}}
\providecommand{\theHalgorithm}{\arabic{algorithm}}
\renewcommand{\theHalgorithm}{supp.\arabic{algorithm}}
\renewcommand{\thefigure}{S\arabic{figure}}
\renewcommand{\thetable}{S\arabic{table}}
\renewcommand{\theequation}{S\arabic{equation}}
\renewcommand{\thealgorithm}{S\arabic{algorithm}}

This supplement reports an additional real-world evaluation on R1, followed by OLED's implementation and evaluation protocol, detailed detection and localization results, source-task analyses, input-window and temporal-inference ablations, and qualitative replays.

\section{Real-World Evaluation}
\label{sec:supp_r1}

\subsection{Dataset}

\begin{figure}[!ht]
\centering
\includegraphics[width=\linewidth]{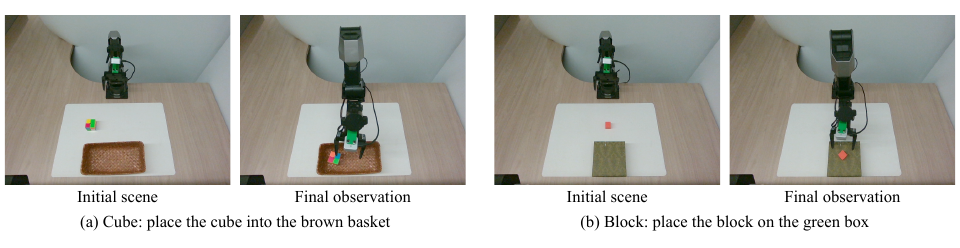}
\caption{The two R1 tasks require placing the cube inside the basket and supporting the block on the box top, respectively.}
\label{fig:supp_r1_tasks}
\end{figure}

We evaluate cross-task transfer on 114 real-world manipulation trajectories collected with a fine-tuned \(\pi_{0.5}\) policy.
The two tasks are placing a cube inside a brown basket (Cube) and placing an orange block on the top surface of a green box (Block).
Fig.~\ref{fig:supp_r1_tasks} illustrates their objects and required placements.
Cube comprises 45 successful and 14 failed trajectories, while Block comprises 36 successful and 19 failed trajectories, yielding 81 successes and 33 failures overall.
Together, these trajectories contain 5,929 front-view RGB observations at \(640\times480\) resolution, with 50--80 observations per trajectory.
The robot operates at 30 Hz, whereas the recorded observations are sampled at approximately 1.52--1.61 Hz for Cube and 0.95--0.99 Hz for Block.
We use the recorded request-reception timestamps to express query intervals and localization errors in seconds.

\subsection{Evaluation Protocol}

\paragraph{Cross-task protocol.}
We evaluate transfer in both directions, Cube-to-Block and Block-to-Cube.
Paired executions from the same scene remain in the same partition, ensuring that fitting and testing groups are disjoint within each fold.
The fitted detector and decision parameters are frozen before evaluation on the corresponding held-out fold of the other task.
Each trajectory is evaluated once in its assigned held-out scene fold, yielding 59 unique target predictions for Cube and 55 for Block.
We compute classification metrics from pooled out-of-fold counts and MAE from the pooled detected failures.
Source BA is computed from held-out source predictions.
Target-task labels are used exclusively for scoring.

\paragraph{Visual inputs and query schedule.}
All detectors receive front-view RGB observations and follow a shared query schedule.
Monitoring begins when the fifth observation becomes available.
Each subsequent query is triggered by the earliest new observation whose timestamp is at least two seconds later than that of the previous query.

\paragraph{Baseline fitting.}
Feature-based baselines use frozen 768-dimensional DINOv2 ViT-B/14 embeddings.
SAFE-MLP and SAFE-LSTM train their respective encoders on source successes and failures.
Similarity fit their anomaly models on source successes and calibrate their thresholds on source validation data.
ViFailback-8B uses a frozen four-image adapter and a source-calibrated threshold.
We evaluate OLED configuration using five-frame JEV assessments.

\begin{table}[t]
\centering
\small
\setlength{\tabcolsep}{5pt}
\caption{Cross-task transfer on the Real-World Evaluation.}
\label{tab:supp_r1_task}
\begin{tabular}{@{}lrrrrrr@{}}
\toprule
Method & Source BA & Target BA & Recall & FPR & MAE (s) & $A_{2\mathrm{s}}$ \\
\midrule
\multicolumn{7}{@{}l}{\textit{Block $\rightarrow$ cube}} \\
\midrule
Similarity & 57.82 & 48.65 & 92.86 & 95.56 & 21.74 & 5.08 \\
SAFE-MLP & 50.07 & 52.54 & 42.86 & 37.78 & 25.52 & 49.15 \\
SAFE-LSTM & 52.56 & 47.62 & 28.57 & 33.33 & 21.57 & 50.85 \\
ViFailback-8B & 37.35 & 53.81 & 14.29 & 6.67 & 5.41 & 71.19 \\
OLED & 60.09 & \textbf{73.89} & 50.00 & 2.22 & 10.57 & 76.27 \\
\midrule
\multicolumn{7}{@{}l}{\textit{Cube $\rightarrow$ Block}} \\
\midrule
Similarity & 85.95 & 44.88 & 84.21 & 94.44 & 30.92 & 3.64 \\
SAFE-MLP & 52.94 & 53.00 & 42.11 & 36.11 & 23.94 & 41.82 \\
SAFE-LSTM & 73.49 & 56.73 & 57.89 & 44.44 & 29.64 & 36.36 \\
ViFailback-8B & 59.21 & 30.77 & 42.11 & 80.56 & 27.45 & 14.55 \\
OLED & 70.56 & \textbf{69.37} & 52.63 & 13.89 & 15.55 & 60.00 \\
\bottomrule
\end{tabular}
\end{table}

\subsection{Evaluation Metrics}

\paragraph{Trajectory-level detection.}
We report source and target BA, together with target Recall, FPR, MAE, and strict accuracy within \(\pm2\) seconds, \(A_{2\mathrm{s}}\).
Failure is the positive class, and a trajectory is predicted as failed if it triggers at least one alarm.
BA averages failure recall and success specificity, \(\mathrm{BA}=\tfrac12(\mathrm{Recall}+1-\mathrm{FPR})\), giving the two outcome classes equal weight.

\paragraph{Strict accuracy.}
Let \(N\) be the number of evaluated trajectories, \(\mathrm{TN}\) the number of successful trajectories without an alarm, and \(\mathcal I_{\mathrm{TP}}\) the set of detected failed trajectories.
Writing \(t_{\mathrm{det}}^{(i)}\) and \(t_{\mathrm{ref}}^{(i)}\) for the first-alarm and reference timestamps in seconds, we compute
\begin{equation}
A_{2\mathrm{s}}=\frac{1}{N}\left(\mathrm{TN}+
\sum_{i\in\mathcal I_{\mathrm{TP}}}
\mathbf{1}\!\left[|t_{\mathrm{det}}^{(i)}-t_{\mathrm{ref}}^{(i)}|\leq2\,\mathrm{s}\right]\right).
\end{equation}
A successful trajectory is correct only if no alarm is issued, while a failed trajectory is correct only if its first alarm falls within the reference window.
Missed failures and first alarms outside the window are counted as incorrect.
This metric jointly rewards correct acceptance of successful executions and first alarms close to the reference time on failed executions.

\paragraph{Timing error.}
We report mean absolute error over detected failures,
\begin{equation}
\mathrm{MAE}=\frac{1}{|\mathcal I_{\mathrm{TP}}|}
\sum_{i\in\mathcal I_{\mathrm{TP}}}
|t_{\mathrm{det}}^{(i)}-t_{\mathrm{ref}}^{(i)}|.
\end{equation}
MAE is measured in seconds and is undefined when no failure is detected.
Its interpretation depends on Recall because each method's detected failures determine the trajectories included in the average.

\subsection{Results and Analysis}

\paragraph{Cross-task detection.}
Table~\ref{tab:supp_r1_task} reports the highest target BA for OLED in both transfer directions.
On Cube, OLED has higher recall and lower FPR than ViFailback-8B, so both terms contribute to its higher BA.
On Block, OLED's recall of \(52.63\%\) is below SAFE-LSTM's \(57.89\%\), but its FPR is also lower, at \(13.89\%\) versus \(44.44\%\).
The larger difference in FPR offsets the recall deficit, giving OLED a BA of \(69.37\%\) compared with \(56.73\%\) for SAFE-LSTM.
The source and target rankings also differ substantially.
For Cube-to-Block transfer, Similarity achieves \(85.95\%\) source BA but only \(44.88\%\) target BA, with a target FPR of \(94.44\%\).
For Similarity in this direction, the higher source BA is accompanied by low target BA and high target FPR.
OLED has the lowest target FPR among the listed methods in both directions, while its recall remains at \(50.00\%\) on Cube and \(52.63\%\) on Block.

\paragraph{Strict accuracy and temporal precision.}
OLED achieves the highest \(A_{2\mathrm{s}}\) in both directions, exceeding the strongest baseline on this metric by \(5.08\) percentage points on Cube and \(18.18\) points on Block.
The ranking by MAE differs on Cube: ViFailback-8B has a lower MAE of \(5.41\) seconds than OLED's \(10.57\) seconds, but also lower recall and strict accuracy.
Thus, the lowest reported MAE on Cube is accompanied by lower failure coverage rather than the highest strict accuracy.
On Block, OLED has the lowest reported MAE at \(15.55\) seconds.

\FloatBarrier

\section{Evaluation Protocol}
\label{sec:supp-protocol}

This section describes the evaluation cohorts, detector implementations, and scoring procedures for the four datasets in Table~\mainpaperref{tab:exp-main}.
We specify dataset-dependent adaptations alongside the shared trajectory-level and first-alarm definitions.

\subsection{Cohort Construction and Task Splits}
\label{sec:supp-cohorts}

\paragraph{LIBERO-Pro.}
We evaluate 200 simulated \(\pi_{0.5}\) rollouts in LIBERO-Pro, using the ten base tasks of LIBERO-10.
The tasks include multi-object placement, spatially specified destinations, and sequential operations such as placing an object in a drawer or microwave and then closing it.
The rollouts cover five perturbation categories involving objects, initial positions, instruction wording, task goals, and environments.
Each rollout belongs to one perturbation category and retains its base-task identity.
We use ten-fold leave-one-task-out evaluation.
Each fold holds out all 20 rollouts from one base task, including its perturbed variants, and uses the remaining nine base tasks as the source pool.
This grouping keeps every variant of the held-out base task outside the source pool.
Each task contributes ten successful and ten failed executions, giving 100 rollouts of each outcome.

\paragraph{ViFailBack.}
We use a fixed binary-outcome subset of 500 real-world ALOHA trajectories from ViFailBack.
The held-out tasks cover object placement, handover, and drawer manipulation, including transferring a pen to a plate and placing a controller in a drawer.
The 17 source tasks contribute 420 trajectories (40 successes and 380 failures).
The five held-out tasks contribute 80 trajectories, with three successes and 13 failures per task.
The source and target task sets are disjoint.
The numerical baseline protocol further divides the source trajectories into 350 fitting and 70 calibration examples.
The calibration partition contains eight successes and 62 failures.
Windows and reused question--answer examples from the same recording stay in the same partition.
I-FailSense uses all 420 source trajectories for fitting and checkpoint selection and applies its native voting procedure to the held-out tasks.

\paragraph{SAFE-WidowX and SAFE-Franka.}
We use two real-robot rollout cohorts released with SAFE, covering different robot platforms and VLA policies.
SAFE-WidowX contains 532 OpenVLA executions across eight tasks, comprising 244 successes and 288 failures.
Its task set consists of four lifting tasks and four object-placement tasks.
SAFE-Franka uses \(\pi_0\)-FAST-DROID executions and includes object placement, stacking, articulated-object manipulation, and cloth unfolding.
The fixed Franka cohort contains 30 successes and 30 failures from each of 13 tasks, giving 780 trajectories with 390 successes and 390 failures.
We use four-fold grouped task holdout separately for each robot.
The WidowX folds hold out two tasks each, with 141, 147, 123, and 121 trajectories.
The Franka folds hold out three, three, three, and four tasks, corresponding to 180, 180, 180, and 240 test trajectories.
The remaining task groups supply the source data in each fold.
The two robots use separate source pools for rule induction, fitting, and calibration.

\subsection{Detector Implementations}
\label{sec:supp-baseline-implementations}

\subsubsection{Baseline Models and Dataset Adaptations}
\label{sec:supp-numerical-scores}

\paragraph{Random Network Distillation (RND).}
RND fits a predictor \(f_\theta\) to the output of a fixed randomly initialized network \(f_{\mathrm{rnd}}\) on a visual embedding \(v_t\).
The prediction error \(\lVert f_\theta(v_t)-f_{\mathrm{rnd}}(v_t)\rVert_2^2\) serves as a novelty score.
For LIBERO-Pro and the two SAFE datasets, the predictor is trained on successful source executions, so a large prediction error indicates a poor match to successful observations.
The ViFailBack adapter fits separate predictors to successful and failed source executions and uses a contrast between their errors as the risk score.
Both outcome classes therefore contribute to the ViFailBack RND score.

\paragraph{LogpZO.}
The RGB adaptation of LogpZO models the distribution of source visual embeddings with a flow-based density model.
For LIBERO-Pro and SAFE, lower estimated density under successful source observations gives higher risk.
For ViFailBack, PCA is fitted on the source fitting set to reduce the embeddings to 20 dimensions.
Separate flow models are then fitted to successful and failed source embeddings, and their density estimates are contrasted for detection.
The fitted projection and density models are reused on held-out queries.

\paragraph{Similarity.}
On LIBERO-Pro and SAFE, the adapter fits PCA and \(k\)-means to successful source embeddings and uses distance to the fitted clusters as its dissimilarity score.
The ViFailBack adapter maintains separate successful and failed source reference sets.
It retrieves up to ten nearest neighbors per outcome using cosine distance and compares the resulting distances.
An embedding farther from successful references relative to failed references receives higher risk.
All reference embeddings and fitted transforms come from the source fitting data.

\paragraph{SAFE.}
Both RGB adaptations of SAFE use frozen DINOv2 visual features and train failure predictors on successful and failed source trajectories.
The two variants differ in their encoding and aggregation of the observed feature history.
SAFE-MLP applies a multilayer perceptron to each feature and uses the running mean of its scalar risk scores to summarize the prefix.
SAFE-LSTM encodes the chronological feature sequence through a recurrent state and maps the current state to a failure score.
On LIBERO-Pro and the SAFE datasets, both variants use the periodically sampled frame-feature history.
On ViFailBack, we extract visual features from five frames sampled within each causal query window and average them into a single feature vector.
These query-level vectors are ordered chronologically to form the observed feature history.
Each variant applies the source-calibrated threshold or time-dependent band specified for the dataset.

\paragraph{I-FailSense.}
I-FailSense combines a frozen vision-language backbone with three lightweight failure-classification heads.
Its adapters uniformly sample four RGB images from the available trajectory prefix and arrange them as a horizontal montage.
The archived LIBERO-Pro training protocol fits the heads on the 180 labeled rollouts from the other nine task families in each fold.
The ViFailBack heads are fitted and selected on all 420 trajectories from the 17 source tasks.
At each query, a valid VLM answer contributes two votes, and each classification head contributes one vote.
The majority of these five votes determines the decision.
When the VLM answer is invalid, the majority of the three heads determines the decision.
On SAFE, the adapter uses the released CALVIN-adapted PaliGemma2-3B checkpoint and initializes the three heads from the released FS weights.
The backbone is frozen while the heads are fitted on terminal four-frame montages from source trajectories.
Training runs for up to 20 epochs, with checkpoint selection followed by evaluation on each causal prefix.
The SAFE threshold is calibrated using the maximum risk over periodic queries on each source calibration trajectory.

\paragraph{JEV direct.}
On ViFailBack and SAFE, the frozen Jev model answers a direct detection question using causal RGB observations and task context.
The ViFailBack request includes the task instruction, objects, goal relations, and the available motion summary.
Its labels are \emph{risk}, \emph{clear}, and \emph{uncertain}, with valid ongoing execution assigned to \emph{clear}.
The recorded ViFailBack detector alarms when the selected label is \emph{risk} and its probability is at least \(0.60\).
The SAFE requests use task context and up to five recent images spanning at most four seconds.
Their risk-probability thresholds are selected on source data from \(0.60,0.65,\ldots,0.95\), with an additional setting that emits no alarm.
The per-query direct decisions are aggregated using the trajectory readout.

\paragraph{ViFailback-8B.}
We use the frozen ViFailback-8B model with its native visual encoder.
At each query, the adapter adds the frames from the current native four-frame window to the set of previously observed frames and removes duplicates.
If this accumulated set contains more than 15 frames, it uniformly subsamples 15 frames while retaining the first and current frames.
The selected images are presented to the model in chronological order.
The model receives these images, their relative temporal offsets, and the task text, and generates a complete answer using its original chat template.
The final answer is parsed as \emph{Failed}, \emph{Successful}, or unknown.
The prompt permits valid ongoing manipulation, and a \emph{Successful} answer indicates that the visible prefix is consistent with valid execution.
A temporal confirmation rule selected on source fitting and validation trajectories converts these query-level labels into an alarm.
The weights remain frozen throughout this selection.

\subsubsection{Visual Inputs, Sampling, and Calibration}
\label{sec:supp-input-interface}
\label{sec:supp-observation}
\label{sec:supp-numerical-fitting}

Each query uses images and task information available by its endpoint.
The implementations differ in image sampling, temporal context, visual encoding, and source calibration, as detailed below.

\paragraph{OLED visual context.}
\Cref{tab:supp_visual_context} lists OLED's recent observations and earlier reference frames relative to the query endpoint.
On LIBERO-Pro and ViFailBack, a query with sufficient history uses 13 recent images and two earlier reference images.
Both SAFE datasets use up to five recent images.
When the available history is shorter, OLED uses the available observations and omits unavailable reference frames.

\begin{table}[!ht]
  \centering
  \small
  \setlength{\tabcolsep}{6pt}
  \caption{OLED's visual inputs at a query with sufficient history.}
  \label{tab:supp_visual_context}
  \begin{tabular}{@{}lccc@{}}
    \toprule
    Cohort & Recent & Span & Reference offsets \\
    \midrule
    LIBERO-Pro & 13 & \(3\,\mathrm{s}\) & \(4\) and \(6\,\mathrm{s}\) \\
    ViFailBack & 13 & \(75\) frames & \(100\) and \(150\) frames \\
    SAFE-WidowX & 5 & \(4\,\mathrm{s}\) & None \\
    SAFE-Franka & 5 & \(4\,\mathrm{s}\) & None \\
    \bottomrule
  \end{tabular}
\end{table}

\paragraph{Visual feature extraction.}
RND, LogpZO, Similarity, SAFE-MLP, and SAFE-LSTM use a frozen DINOv2-Base encoder to extract 768-dimensional visual features in FP32.
On LIBERO-Pro, the adapter averages patch features after resizing the shorter image edge to 256 pixels and center-cropping to \(224\times224\) pixels.
It updates the feature history every five source frames and at the final frame.
On ViFailBack, the adapter samples distinct available frames from \([\max(0,t-75),t]\), where \(t\) is the query endpoint in source frames.
Each image is resized with padding to preserve its full field of view and encoded at \(224\times224\) pixels.
The five feature vectors are averaged into one embedding for the current query.
SAFE-LSTM processes the chronological sequence of these query embeddings.
On SAFE-WidowX and SAFE-Franka, the numerical adapters center-crop images to \(224\times224\) pixels and update features every half-second and at the final frame.
The encoder is frozen while the baseline-specific models are fitted on the source features.

\paragraph{Source calibration.}
\label{sec:supp-calibration}
The numerical baselines determine their alarm thresholds from source calibration trajectories before held-out evaluation.
On LIBERO-Pro, each fold constructs time-dependent calibration bands from successful source trajectories with nominal level \(\alpha=0.05\).
On ViFailBack, calibration first takes the maximum query risk on each of the eight successful source calibration trajectories.
The threshold is the seventh-smallest of these eight maxima, and an alarm requires a risk score strictly above that threshold.
On SAFE-WidowX and SAFE-Franka, each method and fold uses the maximum risk over periodic queries on each source calibration trajectory.
The selected threshold maximizes source calibration BA among operating points with an empirical FPR of at most \(10\%\).

\subsection{Labels, Metrics, and First-Alarm Readout}
\label{sec:supp-evaluation-definitions}

Following \mainpapercref{sec:setup}, we use \(\mathcal D\) for the detector, \(\mathcal Q\) for the query set, \(R_t\) for the binary alarm at query \(t\), and \(t_{\mathrm{det}}\) for the first-alarm index.
The superscript \((i)\) identifies trajectory-specific quantities in the dataset-level evaluation.

\subsubsection{Dataset Labels}
\label{sec:supp-labels}

\paragraph{Trajectory outcomes.}
Each evaluation trajectory \(i\) has a binary outcome label \(y_i\in\{0,1\}\), where \(y_i=1\) denotes failure and \(y_i=0\) denotes success.
Failure is the positive class for every dataset and detector.
LIBERO-Pro uses the recorded task-success outcome of each rollout, while ViFailBack and the two SAFE cohorts retain their recorded outcome annotations.
These labels describe the outcome of the complete execution.
A failed trajectory can contain normal early observations, and a successful trajectory can contain a recoverable adverse event.
Trajectory-level scoring uses the recorded outcome, while frame-level interpretation must account for the execution history.

\paragraph{Temporal reference annotations.}
The ViFailBack localization evaluation additionally uses an annotated reference frame \(t_{\mathrm{ref}}^{(i)}\) for each failed trajectory.
The reference annotation is used exclusively for localization scoring.
The reported error measures temporal agreement between the first alarm and this annotated event.

\subsubsection{Comparison Metrics}
\label{sec:supp-metrics}

\paragraph{Trajectory-level detection.}
Let \(\widehat y_i=1\) when a detector reports at least one alarm during trajectory \(i\), and let \(\widehat y_i=0\) when it reports none.
A failed trajectory with an alarm is a true positive (TP), and a failed trajectory without an alarm is a false negative (FN).
A successful trajectory with an alarm is a false positive (FP), and a successful trajectory without an alarm is a true negative (TN).
An alarm on an execution that later recovers still counts as an FP when its recorded outcome is success.
Writing \(N_+=\mathrm{TP}+\mathrm{FN}\) and \(N_-=\mathrm{TN}+\mathrm{FP}\), we compute
\begin{equation}
  \begin{aligned}
    \mathrm{Recall} &= \frac{\mathrm{TP}}{N_+},\\
    \mathrm{FPR} &= \frac{\mathrm{FP}}{N_-},\\
    \mathrm{BA} &= \frac{1}{2}
      \left(\mathrm{Recall}+1-\mathrm{FPR}\right).
  \end{aligned}
  \label{eq:supp_detection_metrics}
\end{equation}
Recall measures the fraction of failed executions detected, and FPR measures the fraction of successful executions incorrectly rejected.
BA gives equal weight to the two outcome classes.
Detection metrics are reported as percentages.

\paragraph{First-alarm localization.}
For ViFailBack, let \(t_{\mathrm{det}}^{(i)}\) denote the first reported alarm frame, and let \(\mathcal I_{\mathrm{TP}}=\{i:y_i=1,\widehat y_i=1\}\) index the detected failed trajectories.
With \(N=N_++N_-\), the localization metrics are
\begin{equation}
  \begin{aligned}
    H_{25} &= \sum_{i\in\mathcal I_{\mathrm{TP}}}
      \mathbf{1}\!\left[
        |t_{\mathrm{det}}^{(i)}-t_{\mathrm{ref}}^{(i)}|\leq25
      \right],\\
    A_{25} &= \frac{H_{25}+\mathrm{TN}}{N},
      \qquad \mathrm{Recall}_{25} = \frac{H_{25}}{N_+},\\
    \mathrm{MAE} &= \frac{1}{|\mathcal I_{\mathrm{TP}}|}
      \sum_{i\in\mathcal I_{\mathrm{TP}}}
        |t_{\mathrm{det}}^{(i)}-t_{\mathrm{ref}}^{(i)}|.
  \end{aligned}
  \label{eq:supp_general_localization_metrics}
\end{equation}
Here \(\mathbf{1}[\cdot]\) is the indicator function, and both boundary frames of the symmetric \(\pm25\)-frame window are included.
Strict window accuracy \(A_{25}\) rewards in-window detections on failed trajectories and no-alarm decisions on successful trajectories.
Window recall \(\mathrm{Recall}_{25}\) measures temporal hits among all failed trajectories.
Missed failures remain in the denominators of \(A_{25}\) and \(\mathrm{Recall}_{25}\) and contribute zero hits.
MAE is measured in source frames over each method's detected failures, whose membership can differ across methods.
When no failed trajectory is detected, MAE is undefined.

\paragraph{Pooling, repeats, and latency.}
Within each dataset and run, we pool unique held-out predictions across folds to compute classification counts and rates.
Localization MAE is computed over the pooled detected failures.
For repeated runs, each run's metrics are computed first and then averaged with equal weight.
The numerical baselines use seeds 2, 3, and 5 on LIBERO-Pro and 42, 43, and 44 on ViFailBack and SAFE.
Where request latency is reported, P50 and P95 denote its empirical 50th and 95th percentiles, respectively.

\subsubsection{First-Alarm Computation}
\label{sec:supp-first-alarm}

Let \(\mathcal Q^{(i)}\) be the ordered set of scheduled query endpoints for trajectory \(i\), expressed as source-frame indices.
At endpoint \(t\), let \(R_t^{(i)}\in\{0,1\}\) be the trajectory-indexed version of the main paper's alarm \(R_t\), after the detector's thresholding, voting, temporal confirmation, or deferral rule has been applied.
Using the convention \(\min\varnothing=\infty\), we define
\begin{equation}
  \begin{aligned}
    t_{\mathrm{det}}^{(i)} &=
      \min\{t\in\mathcal Q^{(i)}:R_t^{(i)}=1\},\\
    \widehat y_i &= \mathbf{1}[t_{\mathrm{det}}^{(i)}<\infty].
  \end{aligned}
  \label{eq:supp_first_alarm}
\end{equation}
The first alarm is therefore the earliest emitted alarm after the detector's decision procedure has been applied.
A trajectory with no alarm is assigned \(t_{\mathrm{det}}^{(i)}=\infty\).
For a detector with temporal confirmation, the recorded frame is the query where the confirmation condition becomes satisfied.
For example, two consecutive \emph{Failed} answers at frames 25 and 50 yield a first alarm at frame 50 under ViFailback-8B's two-query confirmation setting on ViFailBack.
A deferred OLED candidate is likewise timestamped at the query where its alarm is emitted.
Once recorded, this first alarm remains fixed for both trajectory classification and localization scoring.

\FloatBarrier

\section{Results for Individual Source/Target Splits}
\label{sec:supp-split-results}

Tables~\ref{tab:supp_split_libero_pro}, \ref{tab:supp_split_vifailback}, \ref{tab:supp_split_widowx}, and~\ref{tab:supp_split_franka} report dataset-level detection results and results for each held-out task or task group where applicable.
The ViFailBack results use its single fixed source/target split.
OLED achieves the highest dataset-level BA on all four datasets, with different contributions from failure recall and false-alarm control.
The breakdowns identify the tasks and groups responsible for these aggregate results.
Percentages are rounded to two decimal places after pooling and averaging across runs.
Reconstructing aggregates from the rounded entries can introduce small numerical differences.

\begin{table}[!tp]
  \centering
  \fontsize{8.5}{10}\selectfont
  \setlength{\tabcolsep}{2pt}
  \renewcommand{\arraystretch}{1.02}
  \caption{LIBERO-Pro results by source/target split (\%).
  Column T$k$ identifies base task $k$.}
  \label{tab:supp_split_libero_pro}
  \begin{tabularx}{\textwidth}{@{}ll*{10}{>{\centering\arraybackslash}X}rr@{}}
    \toprule
    Method & Metric & T0 & T1 & T2 & T3 & T4 & T5 & T6 & T7 & T8 & T9 & Pooled & Table 3 \\
    \midrule
    \multicolumn{2}{l}{Failures $N_{+,f}$} & 10 & 10 & 10 & 10 & 10 & 10 & 10 & 10 & 10 & 10 & 100 & -- \\
    \multicolumn{2}{l}{Successes $N_{-,f}$} & 10 & 10 & 10 & 10 & 10 & 10 & 10 & 10 & 10 & 10 & 100 & -- \\
    \midrule
    \multirow{3}{*}{RND} & BA $\uparrow$ & 95.00 & 58.33 & 50.00 & 50.00 & 50.00 & 50.00 & 50.00 & 50.00 & 50.00 & 50.00 & 55.33 & 55.33 \\
     & Rec. $\uparrow$ & 100.00 & 100.00 & 100.00 & 100.00 & 100.00 & 100.00 & 100.00 & 100.00 & 100.00 & 100.00 & 100.00 & 100.00 \\
     & FPR $\downarrow$ & 10.00 & 83.33 & 100.00 & 100.00 & 100.00 & 100.00 & 100.00 & 100.00 & 100.00 & 100.00 & 89.33 & 89.33 \\
    \addlinespace[2pt]
    \multirow{3}{*}{LogpZO} & BA $\uparrow$ & 96.67 & 66.67 & 50.00 & 50.00 & 73.33 & 60.00 & 66.67 & 50.00 & 50.00 & 51.67 & 61.50 & 61.50 \\
     & Rec. $\uparrow$ & 100.00 & 100.00 & 100.00 & 100.00 & 100.00 & 100.00 & 100.00 & 100.00 & 100.00 & 100.00 & 100.00 & 100.00 \\
     & FPR $\downarrow$ & 6.67 & 66.67 & 100.00 & 100.00 & 53.33 & 80.00 & 66.67 & 100.00 & 100.00 & 96.67 & 77.00 & 77.00 \\
    \addlinespace[2pt]
    \multirow{3}{*}{Similarity} & BA $\uparrow$ & 100.00 & 100.00 & 50.00 & 50.00 & 55.00 & 100.00 & 100.00 & 50.00 & 95.00 & 55.00 & 75.50 & 75.50 \\
     & Rec. $\uparrow$ & 100.00 & 100.00 & 100.00 & 100.00 & 100.00 & 100.00 & 100.00 & 100.00 & 100.00 & 100.00 & 100.00 & 100.00 \\
     & FPR $\downarrow$ & 0.00 & 0.00 & 100.00 & 100.00 & 90.00 & 0.00 & 0.00 & 100.00 & 10.00 & 90.00 & 49.00 & 49.00 \\
    \addlinespace[2pt]
    \multirow{3}{*}{SAFE-MLP} & BA $\uparrow$ & 48.33 & 43.33 & 51.67 & 61.67 & 61.67 & 60.00 & 60.00 & 48.33 & 50.00 & 50.00 & 53.50 & 53.50 \\
     & Rec. $\uparrow$ & 10.00 & 0.00 & 100.00 & 100.00 & 53.33 & 50.00 & 40.00 & 6.67 & 0.00 & 0.00 & 36.00 & 36.00 \\
     & FPR $\downarrow$ & 13.33 & 13.33 & 96.67 & 76.67 & 30.00 & 30.00 & 20.00 & 10.00 & 0.00 & 0.00 & 29.00 & 29.00 \\
    \addlinespace[2pt]
    \multirow{3}{*}{SAFE-LSTM} & BA $\uparrow$ & 55.00 & 43.33 & 63.33 & 56.67 & 50.00 & 51.67 & 50.00 & 51.67 & 65.00 & 51.67 & 53.83 & 53.83 \\
     & Rec. $\uparrow$ & 10.00 & 10.00 & 96.67 & 50.00 & 20.00 & 3.33 & 0.00 & 6.67 & 30.00 & 13.33 & 24.00 & 24.00 \\
     & FPR $\downarrow$ & 0.00 & 23.33 & 70.00 & 36.67 & 20.00 & 0.00 & 0.00 & 3.33 & 0.00 & 10.00 & 16.33 & 16.33 \\
    \addlinespace[2pt]
    \multirow{3}{*}{I-FailSense} & BA $\uparrow$ & 60.00 & 65.00 & 60.00 & 55.00 & 50.00 & 50.00 & 50.00 & 40.00 & 60.00 & 55.00 & 54.50 & 54.50 \\
     & Rec. $\uparrow$ & 100.00 & 100.00 & 60.00 & 60.00 & 100.00 & 50.00 & 90.00 & 30.00 & 70.00 & 100.00 & 76.00 & 76.00 \\
     & FPR $\downarrow$ & 80.00 & 70.00 & 40.00 & 50.00 & 100.00 & 50.00 & 90.00 & 50.00 & 50.00 & 90.00 & 67.00 & 67.00 \\
    \addlinespace[2pt]
    \multirow{3}{*}{JEV direct} & BA $\uparrow$ & 60.00 & 50.00 & 50.00 & 50.00 & 50.00 & 50.00 & 50.00 & 55.00 & 50.00 & 50.00 & 51.50 & 51.50 \\
     & Rec. $\uparrow$ & 20.00 & 0.00 & 0.00 & 0.00 & 0.00 & 0.00 & 0.00 & 10.00 & 0.00 & 0.00 & 3.00 & 3.00 \\
     & FPR $\downarrow$ & 0.00 & 0.00 & 0.00 & 0.00 & 0.00 & 0.00 & 0.00 & 0.00 & 0.00 & 0.00 & 0.00 & 0.00 \\
    \addlinespace[2pt]
    \multirow{3}{*}{ViFailback-8B} & BA $\uparrow$ & 60.00 & 80.00 & 95.00 & 95.00 & 90.00 & 95.00 & 85.00 & 55.00 & 55.00 & 85.00 & 79.50 & 79.50 \\
     & Rec. $\uparrow$ & 20.00 & 60.00 & 90.00 & 100.00 & 90.00 & 90.00 & 70.00 & 100.00 & 100.00 & 70.00 & 79.00 & 79.00 \\
     & FPR $\downarrow$ & 0.00 & 0.00 & 0.00 & 10.00 & 10.00 & 0.00 & 0.00 & 90.00 & 90.00 & 0.00 & 20.00 & 20.00 \\
    \midrule
    \multirow{3}{*}{\textbf{OLED}} & BA $\uparrow$ & 100.00 & 85.00 & 85.00 & 100.00 & 80.00 & 80.00 & 90.00 & 85.00 & 60.00 & 65.00 & 83.00 & 83.00 \\
     & Rec. $\uparrow$ & 100.00 & 100.00 & 90.00 & 100.00 & 60.00 & 80.00 & 80.00 & 80.00 & 40.00 & 30.00 & 76.00 & 76.00 \\
     & FPR $\downarrow$ & 0.00 & 30.00 & 20.00 & 0.00 & 0.00 & 20.00 & 0.00 & 10.00 & 20.00 & 0.00 & 10.00 & 10.00 \\
    \bottomrule
  \end{tabularx}
\end{table}

\begin{table}[!tp]
  \centering
  \small
  \setlength{\tabcolsep}{14pt}
  \renewcommand{\arraystretch}{1.08}
  \caption{ViFailBack results for its single fixed source/target split (\%).}
  \label{tab:supp_split_vifailback}
  \begin{tabular}{@{}lrrr@{}}
    \toprule
    Method & BA $\uparrow$ & Rec. $\uparrow$ & FPR $\downarrow$ \\
    \midrule
    RND & 47.52 & 26.15 & 31.11 \\
    LogpZO & 38.46 & 50.26 & 73.33 \\
    Similarity & 49.49 & 12.31 & 13.33 \\
    SAFE-MLP & 48.89 & 0.00 & 2.22 \\
    SAFE-LSTM & 48.89 & 6.67 & 8.89 \\
    I-FailSense & 50.00 & 100.00 & 100.00 \\
    JEV direct & 50.51 & 7.69 & 6.67 \\
    ViFailback-8B & 64.10 & 61.54 & 33.33 \\
    \midrule
    \textbf{OLED} & 71.03 & 75.38 & 33.33 \\
    \bottomrule
  \end{tabular}
\end{table}

\begin{table}[!tp]
  \centering
  \small
  \setlength{\tabcolsep}{2pt}
  \renewcommand{\arraystretch}{1.02}
  \caption{SAFE-WidowX results by source/target split (\%).
  Each column evaluates one held-out task group using the other three groups as sources.}
  \label{tab:supp_split_widowx}
  \begin{tabularx}{\textwidth}{@{}ll*{4}{>{\centering\arraybackslash}X}rr@{}}
    \toprule
    Method & Metric & G1 & G2 & G3 & G4 & Pooled & Table 3 \\
    \midrule
    \multicolumn{2}{l}{Failures $N_{+,f}$} & 79 & 86 & 63 & 60 & 288 & -- \\
    \multicolumn{2}{l}{Successes $N_{-,f}$} & 62 & 61 & 60 & 61 & 244 & -- \\
    \midrule
    \multirow{3}{*}{RND} & BA $\uparrow$ & 58.62 & 50.54 & 48.77 & 41.36 & 49.87 & 49.87 \\
     & Rec. $\uparrow$ & 37.13 & 45.35 & 38.10 & 45.56 & 41.55 & 41.55 \\
     & FPR $\downarrow$ & 19.89 & 44.26 & 40.56 & 62.84 & 41.80 & 41.80 \\
    \addlinespace[2pt]
    \multirow{3}{*}{LogpZO} & BA $\uparrow$ & 51.43 & 42.38 & 46.18 & 40.49 & 45.70 & 45.70 \\
     & Rec. $\uparrow$ & 18.99 & 25.19 & 8.47 & 6.11 & 15.86 & 15.86 \\
     & FPR $\downarrow$ & 16.13 & 40.44 & 16.11 & 25.14 & 24.45 & 24.45 \\
    \addlinespace[2pt]
    \multirow{3}{*}{Similarity} & BA $\uparrow$ & 58.61 & 44.48 & 50.67 & 38.51 & 47.36 & 47.36 \\
     & Rec. $\uparrow$ & 60.76 & 28.29 & 50.79 & 65.00 & 49.77 & 49.77 \\
     & FPR $\downarrow$ & 43.55 & 39.34 & 49.44 & 87.98 & 55.05 & 55.05 \\
    \addlinespace[2pt]
    \multirow{3}{*}{SAFE-MLP} & BA $\uparrow$ & 48.50 & 50.23 & 50.00 & 50.61 & 49.97 & 49.97 \\
     & Rec. $\uparrow$ & 65.82 & 13.57 & 0.00 & 41.11 & 30.67 & 30.67 \\
     & FPR $\downarrow$ & 68.82 & 13.11 & 0.00 & 39.89 & 30.74 & 30.74 \\
    \addlinespace[2pt]
    \multirow{3}{*}{SAFE-LSTM} & BA $\uparrow$ & 54.21 & 53.41 & 52.91 & 53.27 & 53.08 & 53.08 \\
     & Rec. $\uparrow$ & 20.25 & 7.36 & 5.82 & 32.78 & 15.86 & 15.86 \\
     & FPR $\downarrow$ & 11.83 & 0.55 & 0.00 & 26.23 & 9.70 & 9.70 \\
    \addlinespace[2pt]
    \multirow{3}{*}{I-FailSense} & BA $\uparrow$ & 50.00 & 52.53 & 52.30 & 46.32 & 50.21 & 50.21 \\
     & Rec. $\uparrow$ & 5.91 & 14.34 & 7.94 & 17.78 & 11.34 & 11.34 \\
     & FPR $\downarrow$ & 5.91 & 9.29 & 3.33 & 25.14 & 10.93 & 10.93 \\
    \addlinespace[2pt]
    \multirow{3}{*}{JEV direct} & BA $\uparrow$ & 50.00 & 50.00 & 50.00 & 50.00 & 50.00 & 50.00 \\
     & Rec. $\uparrow$ & 0.00 & 0.00 & 0.00 & 0.00 & 0.00 & 0.00 \\
     & FPR $\downarrow$ & 0.00 & 0.00 & 0.00 & 0.00 & 0.00 & 0.00 \\
    \addlinespace[2pt]
    \multirow{3}{*}{ViFailback-8B} & BA $\uparrow$ & 58.05 & 60.57 & 60.16 & 61.02 & 59.41 & 59.41 \\
     & Rec. $\uparrow$ & 17.72 & 24.42 & 26.98 & 45.00 & 27.43 & 27.43 \\
     & FPR $\downarrow$ & 1.61 & 3.28 & 6.67 & 22.95 & 8.61 & 8.61 \\
    \midrule
    \multirow{3}{*}{\textbf{OLED}} & BA $\uparrow$ & 69.66 & 76.87 & 66.94 & 62.84 & 69.70 & 69.70 \\
     & Rec. $\uparrow$ & 68.35 & 84.88 & 55.56 & 66.67 & 70.14 & 70.14 \\
     & FPR $\downarrow$ & 29.03 & 31.15 & 21.67 & 40.98 & 30.74 & 30.74 \\
    \bottomrule
  \end{tabularx}
  \par\vspace{3pt}
  {\footnotesize\raggedright
  The held-out task IDs are G1=$\{0,4\}$, G2=$\{1,5\}$, G3=$\{2,6\}$, and G4=$\{3,7\}$, with 141, 147, 123, and 121 trajectories, respectively.
  \par}
\end{table}

\begin{table}[!ht]
  \centering
  \small
  \caption{First-alarm localization on ViFailBack ($A_{25}$ in \%, MAE in source frames).}
  \label{tab:supp_signed_localization}
  \begin{tabular*}{\linewidth}{@{\extracolsep{\fill}}lrr@{}}
    \toprule
    Method & $A_{25}$ $\uparrow$ & MAE $\downarrow$ \\
    \midrule
    RND& 17.50 & 71.87 \\
    Similarity & 18.75 & 65.63 \\
    LogpZO & 16.25 & 60.96 \\
    I-FailSense & 16.25 & 92.00 \\
    JEV direct & 17.50 & 104.40 \\
    ViFailback-8B & 23.75 & 67.35 \\
    \midrule
    \textbf{OLED} & \textbf{38.75} & \textbf{57.83} \\
    \bottomrule
  \end{tabular*}
\end{table}

\begin{table}[!tp]
  \centering
  \small
  \setlength{\tabcolsep}{2pt}
  \renewcommand{\arraystretch}{1.02}
  \caption{SAFE-Franka results by source/target split (\%).
  Each column evaluates one held-out task group using the other three groups as sources.}
  \label{tab:supp_split_franka}
  \begin{tabularx}{\textwidth}{@{}ll*{4}{>{\centering\arraybackslash}X}rr@{}}
    \toprule
    Method & Metric & G1 & G2 & G3 & G4 & Pooled & Table 3 \\
    \midrule
    \multicolumn{2}{l}{Failures $N_{+,f}$} & 90 & 90 & 90 & 120 & 390 & -- \\
    \multicolumn{2}{l}{Successes $N_{-,f}$} & 90 & 90 & 90 & 120 & 390 & -- \\
    \midrule
    \multirow{3}{*}{RND} & BA $\uparrow$ & 49.63 & 50.19 & 47.04 & 46.25 & 48.12 & 48.12 \\
     & Rec. $\uparrow$ & 64.44 & 36.30 & 27.41 & 18.06 & 35.13 & 35.13 \\
     & FPR $\downarrow$ & 65.19 & 35.93 & 33.33 & 25.56 & 38.89 & 38.89 \\
    \addlinespace[2pt]
    \multirow{3}{*}{LogpZO} & BA $\uparrow$ & 48.70 & 46.11 & 45.19 & 45.83 & 46.41 & 46.41 \\
     & Rec. $\uparrow$ & 67.78 & 28.89 & 44.44 & 12.22 & 36.32 & 36.32 \\
     & FPR $\downarrow$ & 70.37 & 36.67 & 54.07 & 20.56 & 43.50 & 43.50 \\
    \addlinespace[2pt]
    \multirow{3}{*}{Similarity} & BA $\uparrow$ & 46.67 & 51.67 & 47.04 & 50.00 & 48.93 & 48.93 \\
     & Rec. $\uparrow$ & 23.33 & 8.15 & 11.11 & 25.56 & 17.69 & 17.69 \\
     & FPR $\downarrow$ & 30.00 & 4.81 & 17.04 & 25.56 & 19.83 & 19.83 \\
    \addlinespace[2pt]
    \multirow{3}{*}{SAFE-MLP} & BA $\uparrow$ & 55.93 & 54.44 & 49.81 & 50.00 & 52.35 & 52.35 \\
     & Rec. $\uparrow$ & 14.44 & 8.89 & 1.11 & 3.89 & 6.84 & 6.84 \\
     & FPR $\downarrow$ & 2.59 & 0.00 & 1.48 & 3.89 & 2.14 & 2.14 \\
    \addlinespace[2pt]
    \multirow{3}{*}{SAFE-LSTM} & BA $\uparrow$ & 52.96 & 50.37 & 51.85 & 50.97 & 51.50 & 51.50 \\
     & Rec. $\uparrow$ & 30.74 & 0.74 & 5.56 & 24.72 & 16.15 & 16.15 \\
     & FPR $\downarrow$ & 24.81 & 0.00 & 1.85 & 22.78 & 13.16 & 13.16 \\
    \addlinespace[2pt]
    \multirow{3}{*}{I-FailSense} & BA $\uparrow$ & 48.33 & 52.59 & 50.74 & 45.42 & 48.97 & 48.97 \\
     & Rec. $\uparrow$ & 11.85 & 6.30 & 22.96 & 18.33 & 15.13 & 15.13 \\
     & FPR $\downarrow$ & 15.19 & 1.11 & 21.48 & 27.50 & 17.18 & 17.18 \\
    \addlinespace[2pt]
    \multirow{3}{*}{JEV direct} & BA $\uparrow$ & 50.00 & 51.11 & 51.11 & 49.58 & 50.38 & 50.38 \\
     & Rec. $\uparrow$ & 0.00 & 2.22 & 2.22 & 0.00 & 1.03 & 1.03 \\
     & FPR $\downarrow$ & 0.00 & 0.00 & 0.00 & 0.83 & 0.26 & 0.26 \\
    \addlinespace[2pt]
    \multirow{3}{*}{ViFailback-8B} & BA $\uparrow$ & 52.22 & 51.11 & 51.67 & 49.58 & 51.03 & 51.03 \\
     & Rec. $\uparrow$ & 4.44 & 2.22 & 13.33 & 25.83 & 12.56 & 12.56 \\
     & FPR $\downarrow$ & 0.00 & 0.00 & 10.00 & 26.67 & 10.51 & 10.51 \\
    \midrule
    \multirow{3}{*}{\textbf{OLED}} & BA $\uparrow$ & 51.11 & 74.44 & 51.67 & 42.92 & 54.10 & 54.10 \\
     & Rec. $\uparrow$ & 55.56 & 70.00 & 90.00 & 60.00 & 68.21 & 68.21 \\
     & FPR $\downarrow$ & 53.33 & 21.11 & 86.67 & 74.17 & 60.00 & 60.00 \\
    \bottomrule
  \end{tabularx}
  \par\vspace{3pt}
  {\footnotesize\raggedright
  The held-out task IDs are G1=$\{0,2,12\}$, G2=$\{1,6,7\}$, G3=$\{3,4,11\}$, and G4=$\{5,8,9,10\}$, with 180, 180, 180, and 240 trajectories, respectively.
  \par}
\end{table}

\paragraph{LIBERO-Pro.}
OLED reaches a pooled BA of \(83.00\%\), exceeding ViFailback-8B, the strongest baseline, by \(3.50\) percentage points in Table~\ref{tab:supp_split_libero_pro}.
Compared with ViFailback-8B, its recall decreases from \(79.00\%\) to \(76.00\%\), while its FPR decreases from \(20.00\%\) to \(10.00\%\).
The ten-percentage-point reduction in FPR offsets the three-point reduction in recall, producing the \(3.50\)-point BA gain.
RND, LogpZO, and Similarity all achieve \(100.00\%\) recall, but their FPRs of \(89.33\%\), \(77.00\%\), and \(49.00\%\), respectively, limit their BA.
At the task level, OLED attains \(100.00\%\) BA on T0 and T3 and exceeds ViFailback-8B on six of the ten tasks.
On T7, its BA is \(85.00\%\), compared with \(55.00\%\) for ViFailback-8B, with FPR reduced from \(90.00\%\) to \(10.00\%\).
The task breakdown also reveals limited failure coverage on T8 and T9.
OLED's recall falls to \(40.00\%\) and \(30.00\%\) on these tasks, leaving BA at \(60.00\%\) and \(65.00\%\), respectively.

\paragraph{ViFailBack.}
In Table~\ref{tab:supp_split_vifailback}, OLED improves BA from ViFailback-8B's \(64.10\%\) to \(71.03\%\) at the same \(33.33\%\) FPR.
OLED's recall is \(75.38\%\), compared with \(61.54\%\) for ViFailback-8B.
With FPR unchanged, the BA difference is entirely attributable to the higher reported recall.
For RND, results computed from archived trajectory predictions for seeds 42, 43, and 44, with equal weight per run, are \(47.52\%\) BA, \(26.15\%\) recall, and \(31.11\%\) FPR.
Its FPR exceeds its recall, leaving BA below \(50.00\%\).
OLED has the highest BA, and ViFailback-8B is the strongest baseline.
SAFE-MLP produces few false alarms but detects no failures, yielding \(48.89\%\) BA despite an FPR of \(2.22\%\).
I-FailSense alarms on every trajectory, giving \(100.00\%\) recall and FPR and \(50.00\%\) BA.
These outcomes show how each extreme limits discrimination between successful and failed executions.

\paragraph{SAFE-WidowX.}
Table~\ref{tab:supp_split_widowx} shows a pooled BA of \(69.70\%\) for OLED, exceeding ViFailback-8B's \(59.41\%\) by \(10.29\) percentage points.
OLED ranks first in BA in all four groups, with margins over the strongest baseline in each group ranging from \(1.82\) to \(16.30\) percentage points.
Its strongest result is G2, with \(76.87\%\) BA and \(84.88\%\) recall.
G4 remains the most difficult group for OLED, with \(62.84\%\) BA and its highest group FPR of \(40.98\%\).
Compared with ViFailback-8B at the dataset level, OLED raises recall from \(27.43\%\) to \(70.14\%\), at the cost of increasing FPR from \(8.61\%\) to \(30.74\%\).
The recall increase exceeds the FPR increase, accounting for the higher BA.
The higher pooled BA therefore accompanies a higher FPR, with G4 showing OLED's largest group-level FPR.

\paragraph{SAFE-Franka.}
OLED obtains \(54.10\%\) pooled BA in \cref{tab:supp_split_franka}, compared with \(52.35\%\) for SAFE-MLP, the strongest baseline.
G2 makes the strongest contribution to this \(1.75\)-point advantage.
OLED reaches \(74.44\%\) BA in G2, exceeding the best baseline by \(20.00\) percentage points, while other methods achieve higher BA in each of the remaining groups.
In G3, \(90.00\%\) recall is largely offset by an FPR of \(86.67\%\), leaving BA at \(51.67\%\).
In G4, the FPR of \(74.17\%\) exceeds recall of \(60.00\%\), reducing BA to \(42.92\%\).
The pooled comparison with SAFE-MLP shows a large increase in recall, from \(6.84\%\) to \(68.21\%\), accompanied by an FPR increase from \(2.14\%\) to \(60.00\%\).
In these reported results, the small pooled BA advantage accompanies a substantial FPR increase, and G3 and G4 show the largest FPRs for OLED.

\FloatBarrier

\section{First-Alarm Localization}
\label{sec:supp_localization}

We examine first-alarm timing on the five held-out ViFailBack tasks using the reference annotations and metrics defined in \cref{sec:supp-metrics}.
The tabulated results use an inclusive tolerance of \(\pm25\) source frames.

\paragraph{Aggregate localization results.}
Table~\ref{tab:supp_signed_localization} reports an \(A_{25}\) of \(38.75\%\) for OLED, compared with \(23.75\%\) for ViFailback-8B.
OLED also has the lowest reported MAE among the methods in the table.
However, MAE is conditional on each method's detected failures, whereas \(A_{25}\) evaluates strict correctness across the entire cohort, including successful executions without alarms.
LogpZO illustrates this distinction: its MAE of \(60.96\) frames is close to OLED's \(57.83\) frames, but its \(A_{25}\) is only \(16.25\%\).
A relatively low error among detected failures therefore does not imply accurate detection and localization across the full cohort.
Conversely, RND and JEV direct share an \(A_{25}\) of \(17.50\%\), but their MAEs are \(71.87\) and \(104.40\) frames, respectively.
The equal strict accuracies therefore do not imply equal timing errors among detected failures.

\begin{figure}[!htbp]
  \centering
  \includegraphics[width=\linewidth]{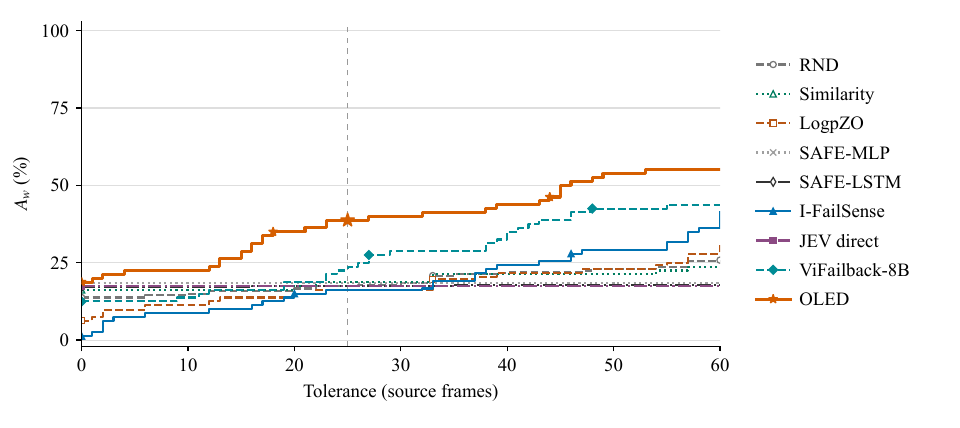}
  \caption{Strict window accuracy over tolerances of 0--60 source frames using archived localization readouts.
  The dashed line marks \(w=25\), where the curves give \(A_{25}\).}
  \label{fig:supp_localization_tolerance}
\end{figure}

\paragraph{Sensitivity to the localization tolerance.}
\Cref{fig:supp_localization_tolerance} varies the tolerance \(w\) from 0 to 60 source frames while retaining each method's recorded first alarms.
Let \(H_w\) count failed trajectories whose first alarm falls within \(\pm w\) frames of the reference.
We report strict window accuracy \(A_w=(H_w+\mathrm{TN})/N\), which weights all 80 trajectories equally and reduces to \(A_{25}\) at \(w=25\).
For the archived localization readouts, OLED reaches \(38.75\%\) at this tolerance, compared with \(23.75\%\) for ViFailback-8B.
Increasing the tolerance to 60 frames raises OLED's \(A_w\) to \(55.00\%\), indicating that some detected failures are localized outside the narrower 25-frame window.
Because each method's true-negative count remains fixed, increases along its curve reflect existing failure alarms admitted by a wider tolerance, rather than additional trajectory-level detections.
Differences between methods jointly reflect temporal agreement on failed trajectories and false-alarm control on successful trajectories.
Missed failures and false alarms on successful trajectories remain incorrect at every tolerance.

\FloatBarrier

\section{Source Task Composition}
\label{sec:supp-source-ablation}

We examine how source rule families guide the generation of monitoring conditions for a fixed target task.
\Cref{tab:supp-case-rule-dictionary} compares the resulting candidates in terms of constraint specificity and execution contexts.
The source sets are nested, with \texttt{task03} and \texttt{task00} in the two-task condition.
The four-task condition adds \texttt{task05} and \texttt{task06}, and the six-task condition further adds \texttt{task08} and \texttt{task02}.
Each source task contributes five static cards, giving 10, 20, and 30 cards in the three conditions.
The zero-source condition generates candidates directly from the target task.

\begin{figure}[!ht]
\centering
\includegraphics[width=0.43\linewidth]
  {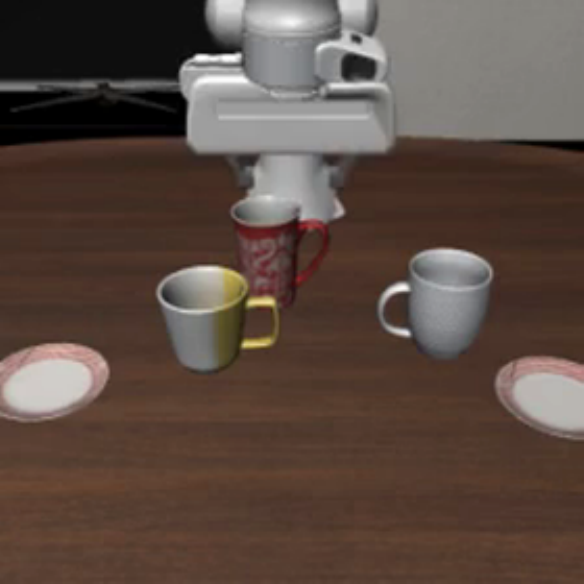}
\caption{Real first RGB frame used for the fixed target-task
example. The task requires the white mug on the task-left plate
and the yellow--white mug on the task-right plate.
The image alone does not certify the correspondence between
task-relative and image-relative directions.}
\label{fig:supp-source-target-scene}
\end{figure}

\begin{table}[t]
\centering
\footnotesize
\setlength{\tabcolsep}{4pt}
\renewcommand{\arraystretch}{1.12}
\caption{Generated rule candidates for the fixed task \emph{place the white mug on the left plate and the yellow--white mug on the right plate}.
Here \(k\) denotes the number of source tasks.
Rule IDs are local to each column.}
\label{tab:supp-case-rule-dictionary}
\begin{tabularx}{\textwidth}
  {@{}l*{4}{>{\raggedright\arraybackslash}X}@{}}
\toprule
Local ID
& $k=0$: no source rules
& $k=2$: source-conditioned
& $k=4$: source-conditioned
& $k=6$: source-conditioned \\
\midrule
r1
& white mug lacks goal-plate and other support after release$^\dagger$
& white mug placed on a non-goal plate
& white mug supported by the wrong plate
& white mug supported by the wrong plate \\
r2
& yellow--white mug lacks goal-plate and other support after release$^\dagger$
& yellow--white mug placed on a non-goal plate
& yellow--white mug supported by the wrong plate
& yellow--white mug supported by the wrong plate \\
r3
& carried mug collision causes displacement or tipping
& uncontrolled free fall or support loss during transport
& grasped mug falls without plate or table support
& carried mug falls without plate or table support \\
r4
& held mug slips or falls from the gripper
& carried mug strikes an obstacle and causes displacement or deformation
& carried mug collision causes displacement or deformation
& placed mug tips or rolls off its plate \\
r5
& released mug is supported by the other mug's required plate
& wrong mug occupies the required plate
& placed mug tips or rolls off its plate
& carried mug collision causes displacement, excluding goal-plate contact \\
r6
& ---
& target mug knocked off the table before grasping
& previously placed mug loses plate support
& non-target object occupies a goal plate while its required mug is absent \\
r7
& ---
& ---
& non-target red mug occupies a goal plate
& previously placed mug loses plate support \\
r8
& ---
& ---
& carried mug contacts and displaces another mug
& --- \\
\bottomrule
\end{tabularx}
\end{table}

\paragraph{More specific target constraints.}
Source conditioning makes the affected objects and applicability conditions more explicit in the listed candidates.
All three positive-source settings give the white and yellow--white mugs separate wrong-plate checks, whereas the zero-source output uses a shared wrong-plate rule alongside two unsupported-release rules.
The six-source output further qualifies its collision rule by excluding goal-plate contact and its occupancy rule by requiring the intended mug to be absent.
Compared with the four-source occupancy rule naming a non-target red mug, the six-source rule covers a broader object category while specifying this additional condition.
These differences show more explicit object bindings and applicability qualifications for the fixed target task.

\paragraph{Additional execution contexts.}
Source conditioning also adds explicitly named pre-grasp and post-placement conditions to the transport and release events in the zero-source list.
The two-source output includes a target mug being knocked off the table before grasping, while the four- and six-source outputs describe a placed mug tipping or rolling off its plate and a previously placed mug losing plate support.
These entries distinguish pre-grasp displacement and loss of an established plate relation from failures during transport or release.
The candidate counts are 5, 6, 8, and 7 across the four settings, with overlap among some post-placement descriptions.
The additional contexts support a richer description of potential events for this task.

\FloatBarrier

\section{Effect of Input Window}
\label{sec:supp-input-window}

\begin{table}[t]
\centering
\small
\setlength{\tabcolsep}{8pt}
\caption{Input-window ablation on LIBERO-Pro with 30 calibration and 20 test trajectories.
Each window is calibrated independently on the same partition.}
\label{tab:supp-input-window}
\begin{tabular*}{\textwidth}{@{\extracolsep{\fill}}ccccc@{}}
\toprule
Window (s) & Images & BA $\uparrow$ & Recall $\uparrow$ & FPR $\downarrow$ \\
\midrule
1 & 5 & $80.49$ & $72.09$ & $11.12$ \\
2 & 9 & $81.67$ & $74.97$ & $11.64$ \\
3 & 13 & $\mathbf{85.88}$ & $\mathbf{79.93}$ & $\mathbf{8.18}$ \\
4 & 17 & $83.94$ & $76.36$ & $8.49$ \\
\bottomrule
\end{tabular*}
\end{table}

\paragraph{Experimental setup.}
We compare causal RGB windows spanning 1, 2, 3, and 4 seconds on a fixed subset of 50 LIBERO-Pro trajectories, with 25 successes and 25 failures.
Images are sampled every five source frames at 20 FPS, giving 5, 9, 13, and 17 images, respectively, when sufficient history is available.
The Jev model, task rules, query endpoints, and candidate configuration grid are held fixed across window lengths.
In each partitions, 30 trajectories (15 successes and 15 failures) are used for calibration and the remaining 20 (ten of each outcome) for testing.
Each window independently selects its temporal and confirmation parameters using calibration data only, and the selected configuration is frozen for testing.
\Cref{tab:supp-input-window} reports mean test metrics across these partitions of the same 50 trajectories.

\paragraph{Effect of temporal context.}
\Cref{tab:supp-input-window} shows that detection quality does not improve monotonically with window duration, even when each window is calibrated independently.
Extending the window from one to two seconds improves recall but also increases false alarms, so the additional failure coverage yields only a modest BA gain.
At three seconds, recall improves further while FPR decreases, producing the highest mean BA (\(85.88\%\)) among the evaluated settings.
The three-second setting also has the highest recall and lowest FPR in the table, so its BA advantage combines improvements in both terms.
Extending the window to four seconds reduces recall by \(3.57\) percentage points and increases FPR by \(0.31\) points, lowering BA by \(1.94\) points relative to three seconds.
The recall decrease is larger than the FPR increase, making it the larger contributor to the BA decline.
Three seconds provides the best observed balance between failure coverage and false alarms on this cohort.
Because longer windows also contain more images, this comparison reflects the combined effect of temporal span and input size.

\FloatBarrier

\section{Detailed Temporal-Inference Ablations}
\label{sec:supp_temporal_ablation_detail}

\begin{table}[t]
\centering
\small
\setlength{\tabcolsep}{6pt}
\caption{Detailed records associated with the temporal-inference ablations.}
\label{tab:supp_temporal_ablation_detail}
\begin{tabular*}{\textwidth}{@{\extracolsep{\fill}}llrrrrrrr@{}}
\toprule
Dataset & Variant & BA $\uparrow$ & Recall $\uparrow$ & FPR $\downarrow$ & TP & FP & FN & TN \\
\midrule
\multirow{4}{*}{LIBERO-Pro}
& w/o goal-route & 76.00 & 56.00 & 4.00 & 56 & 4 & 44 & 96 \\
& w/o event-route & 79.50 & 69.00 & 10.00 & 69 & 10 & 31 & 90 \\
& w/o confirmation & 80.00 & 76.00 & 16.00 & 76 & 16 & 24 & 84 \\
& OLED & \textbf{83.00} & 76.00 & 10.00 & 76 & 10 & 24 & 90 \\
\midrule
\multirow{4}{*}{ViFailBack}
& w/o goal-route & 55.90 & 18.46 & 6.67 & 12 & 1 & 53 & 14 \\
& w/o event-route & 69.49 & 72.31 & 33.33 & 47 & 5 & 18 & 10 \\
& w/o confirmation & 56.41 & 86.15 & 73.33 & 56 & 11 & 9 & 4 \\
& OLED & \textbf{71.03} & 75.38 & 33.33 & 49 & 5 & 16 & 10 \\
\midrule
\multirow{4}{*}{SAFE-WidowX}
& w/o goal-route & 62.41 & 35.07 & 10.25 & 101 & 25 & 187 & 219 \\
& w/o event-route & 66.94 & 62.15 & 28.28 & 179 & 69 & 109 & 175 \\
& w/o confirmation & 67.30 & 80.90 & 46.31 & 233 & 113 & 55 & 131 \\
& OLED & \textbf{69.70} & 70.14 & 30.74 & 202 & 75 & 86 & 169 \\
\midrule
\multirow{4}{*}{SAFE-Franka}
& w/o goal-route & 51.41 & 32.56 & 29.74 & 127 & 116 & 263 & 274 \\
& w/o event-route & 52.95 & 66.15 & 60.26 & 258 & 235 & 132 & 155 \\
& w/o confirmation & 53.21 & 70.77 & 64.36 & 276 & 251 & 114 & 139 \\
& OLED & \textbf{54.10} & 68.21 & 60.00 & 266 & 234 & 124 & 156 \\
\bottomrule
\end{tabular*}
\end{table}

Table~\ref{tab:supp_temporal_ablation_detail} decomposes the reported BA into failure recall and false-positive rate.
OLED has the highest BA among the listed variants on each dataset, although it does not always achieve the highest recall or the lowest FPR.
The variant without the goal route has the lowest recall on all four datasets.
On ViFailBack, it detects only 12 of 65 failures, compared with 49 for OLED; the reduction from five false alarms to one does not offset this loss of failure coverage, and BA falls from \(71.03\%\) to \(55.90\%\).
The event-route comparison shows smaller recall differences: OLED detects seven additional failures on LIBERO-Pro and two on ViFailBack with unchanged false-positive counts.
On SAFE-WidowX, OLED detects 23 additional failures at the cost of six additional false alarms, increasing BA from \(66.94\%\) to \(69.70\%\).

The confirmation comparison shows different recall--FPR trade-offs across the listed datasets.
On LIBERO-Pro, both OLED and the variant without confirmation detect 76 failures, but OLED reduces false alarms from 16 to 10, raising BA from \(80.00\%\) to \(83.00\%\) without reducing recall.
On ViFailBack, removing confirmation detects seven additional failures but produces six additional false alarms.
Because this cohort contains 65 failures and only 15 successful trajectories, these changes correspond to a \(10.77\)-percentage-point recall increase and a \(40.00\)-point FPR increase, reducing BA by \(14.62\) points.
SAFE-WidowX shows a similar trade-off: removing confirmation adds 31 detected failures and 38 false alarms, reducing BA from \(69.70\%\) to \(67.30\%\).
On SAFE-Franka, the corresponding increases are ten detected failures and 17 false alarms, yielding a smaller BA reduction of less than one percentage point.

Among the four datasets in the table, SAFE-Franka has the lowest BA for OLED.
OLED misses 124 of 390 failures and incorrectly flags 234 of 390 successful trajectories.
Its \(68.21\%\) recall and \(60.00\%\) FPR yield only \(54.10\%\) BA, indicating limited separation between successful and failed executions.
These trajectory-level results characterize detection trade-offs.

\FloatBarrier

\section{Qualitative Analysis of Temporal Localization}
\label{sec:supp_qualitative}

We supplement the quantitative evaluation with six failure cases from LIBERO-Pro and SAFE-Franka, illustrating the relation between visible placement violations and first-alarm timing.
Each sequence pairs a task-specific rule with RGB observations and query probabilities to examine how the reported exception relates to the observed execution.
The displayed rules paraphrase the corresponding EVENT questions, and post-alarm images provide retrospective context.

\begin{figure}[p]
  \centering
  \includegraphics[width=\textwidth]{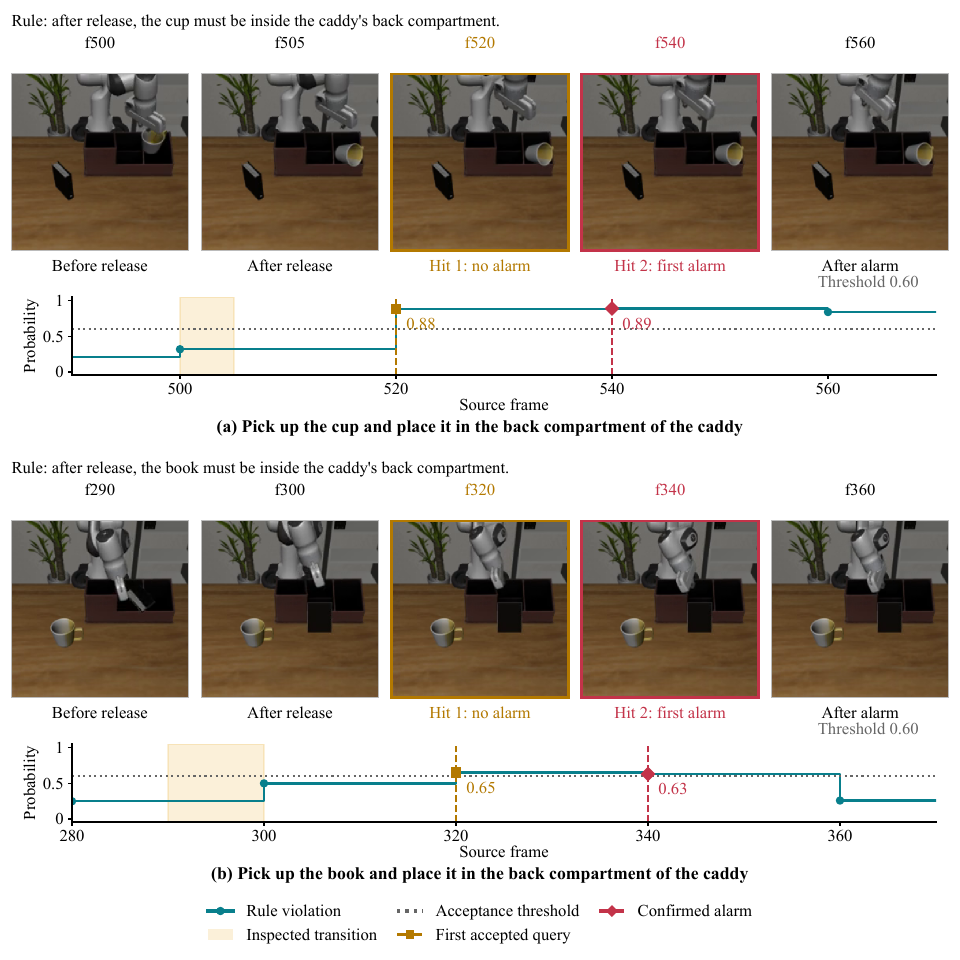}
  \caption{Placement violations in LIBERO-Pro.
  Amber squares indicate the first accepted query; red diamonds mark the confirming query and first alarm.
  Dots denote query probabilities, and step lines hold the preceding value between queries.}
  \label{fig:supp_temporal_cases}
\end{figure}

\begin{figure}[p]
  \centering
  \includegraphics[width=\textwidth]{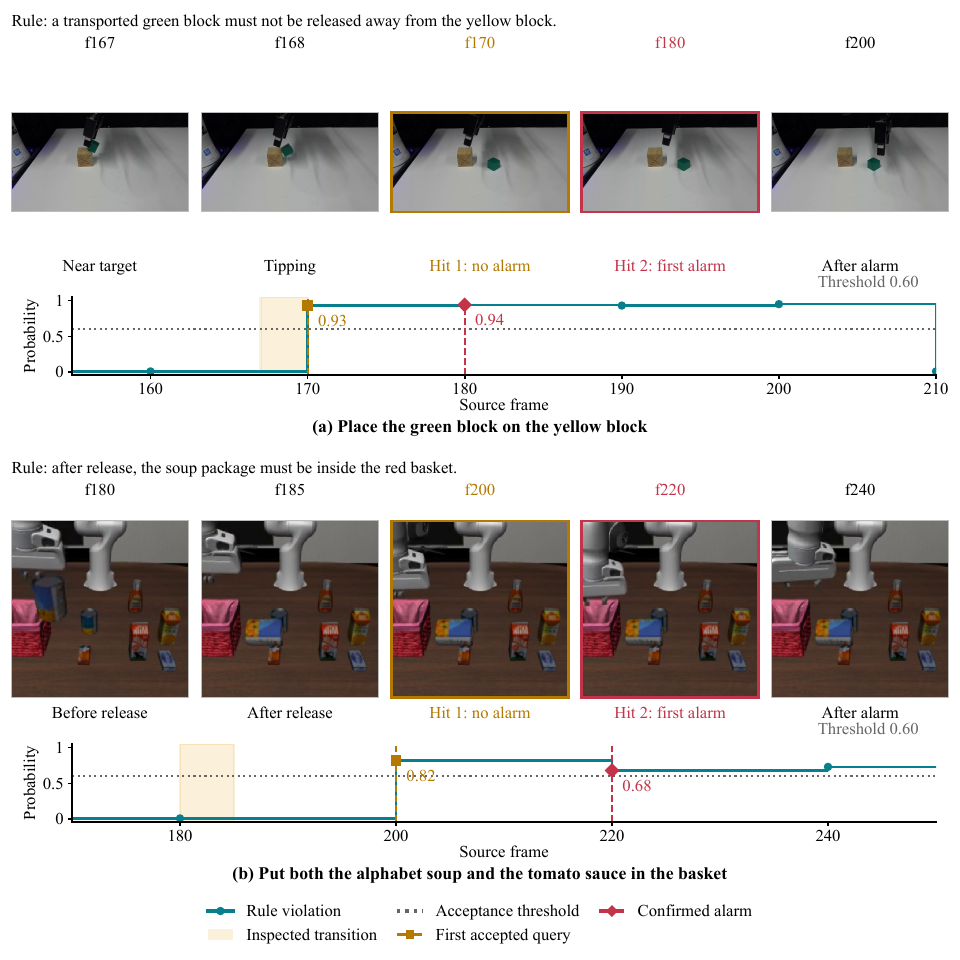}
  \caption{Support and containment violations.
  (a) A block lands beside its target in SAFE-Franka.
  (b) A soup package is released outside the basket in LIBERO-Pro.}
  \label{fig:supp_temporal_additional_a}
\end{figure}

\begin{figure}[p]
  \centering
  \includegraphics[width=\textwidth]{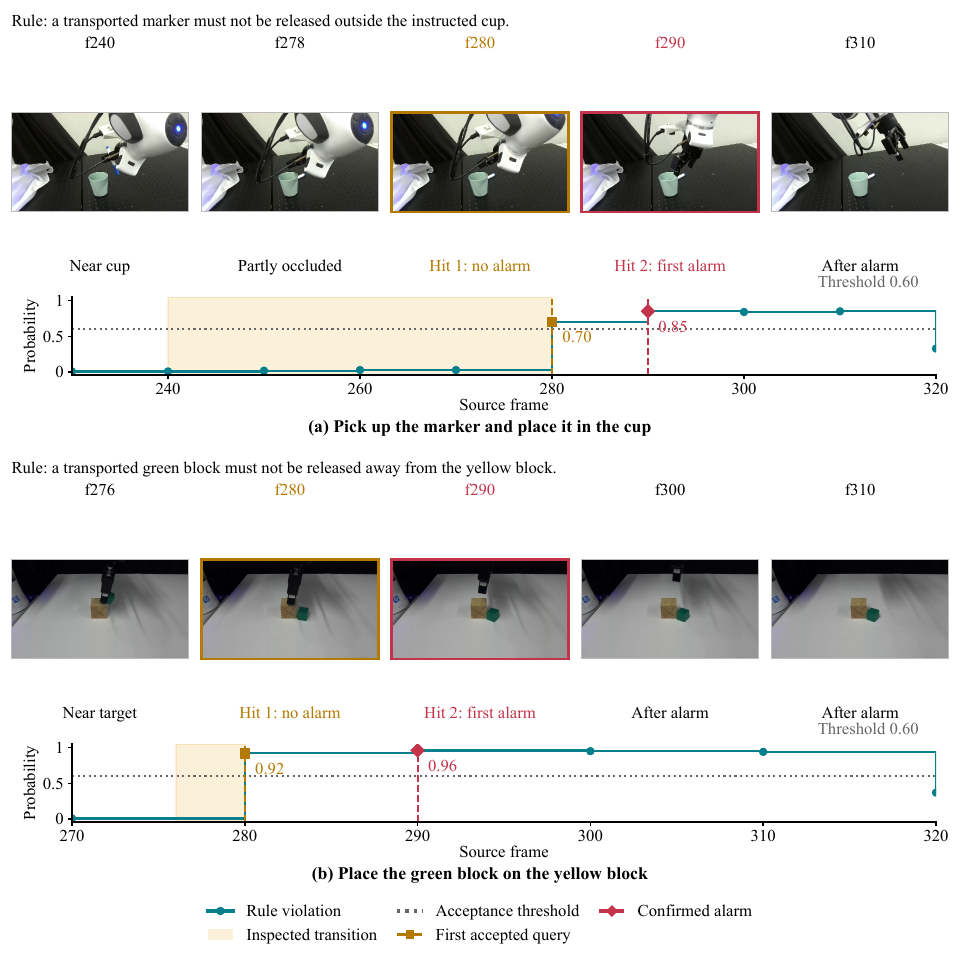}
  \caption{Additional SAFE-Franka failures.
  (a) A marker remains outside the cup after a partly occluded release.
  (b) A second block trajectory ends beside the target.}
  \label{fig:supp_temporal_additional_b}
\end{figure}

Figures~\ref{fig:supp_temporal_cases}--\ref{fig:supp_temporal_additional_b} show alarms associated with persistent violations of the required support or containment relation after object release.
In Fig.~\ref{fig:supp_temporal_cases}, the cup remains at the caddy's outer edge and the book is released onto the table in front of it, providing visible evidence of incompatible placement rather than simply an unfinished task.
For both cases, the first accepted event query leaves \(E_t=0\), whereas the next adjacent accepted query activates event memory and produces the first alarm.
The book's violation probability decreases from \(0.65\) to \(0.63\) across these queries, illustrating that confirmation depends on sustained acceptance rather than increasing confidence.
Figure~\ref{fig:supp_temporal_additional_a} extends the analysis to a real block-placement failure and a simulated soup-package placement failure: the objects remain beside the required support and outside the required container, respectively.
Figure~\ref{fig:supp_temporal_additional_b} shows the same relation-based interpretation for a marker and a second block trajectory.
Although the marker's release is partly occluded, its subsequent outside-cup position supports the placement judgment; the broader inspected interval reflects uncertainty about the exact transition time.
These cases illustrate how task rules provide an interpretable account of an execution exception across different objects and environments.
The alarms follow the observed transitions and include sampling and confirmation delays; the qualitative intervals are not treated as ground-truth localization errors.

\FloatBarrier

\end{document}